\documentclass[letterpaper]{article} 
\usepackage[preprint]{aaai2027}  
\usepackage[hyphens]{url}  
\usepackage{graphicx} 
\usepackage{natbib}  
\usepackage{caption} 
\usepackage{algorithm}
\usepackage{algorithmic}

\usepackage{newfloat}
\usepackage{listings}
\DeclareCaptionStyle{ruled}{labelfont=normalfont,labelsep=colon,strut=off} 
\floatstyle{ruled}
\newfloat{listing}{tb}{lst}{}
\floatname{listing}{Listing}

\usepackage{booktabs}
\usepackage{array}
\usepackage{tabularx}
\usepackage{colortbl}
\usepackage{multirow}
\usepackage{amsmath}
\usepackage{amssymb}
\newcolumntype{Y}{>{\raggedright\arraybackslash}X}
\definecolor{resultheadergray}{gray}{0.78}
\definecolor{resultrowgray}{gray}{0.88}
\newcommand{\figtarget}[1]{\pdfdest name {#1} xyz}
\newcommand{\figlink}[1]{\leavevmode\pdfstartlink attr {/Border [0 0 0]} goto name {#1}Figure~\ref{#1}\pdfendlink}
\newcommand{\tabtarget}[1]{\pdfdest name {#1} xyz}
\newcommand{\tablink}[1]{\leavevmode\pdfstartlink attr {/Border [0 0 0]} goto name {#1}Table~\ref{#1}\pdfendlink}
\newcommand{\sectarget}[1]{\pdfdest name {#1} xyz}

\newcommand{\subseclink}[1]{\leavevmode\pdfstartlink attr {/Border [0 0 0]} goto name {#1}Subsection~\ref{#1}\pdfendlink}

\title{GUARD: \textbf{G}rounding \textbf{U}ncertainty and \textbf{A}blation-Based \textbf{R}isk \textbf{D}etection for Diffusion-Based VLAs}
\author{
    Suhas Hegde\textsuperscript{\rm 1},
    Jitendra Yasaswi Bharadwaj Katta\textsuperscript{\rm 1}
}
\affiliations{
    \textsuperscript{\rm 1}Bosch Research India
}

\begin{document}

\maketitle

\begin{abstract}
Diffusion-based vision-language-action (VLA) policies can generate plausible actions even when their predictions are weakly grounded in the visual and language evidence defining the task. We introduce GUARD, a test-time failure detection method that measures this grounding without modifying the pretrained policy. GUARD estimates the influence of token-indexed entries in the final vision-language model key--value (KV) cache, constructs counterfactual caches by ablating salient KV entries, and compares their denoising responses with the original conditioning. Based on the comparison, we derive GUARD diagnostic stream including \textit{sensitivity}, \textit{attention entropy}, \textit{modality bias}, and \textit{grounding efficiency}, which are calibrated online and processed by a lightweight temporal classifier. We evaluate GUARD under task-held-out splits across five policy--benchmark settings, using Pi0, SmolVLA, and Alpamayo-1.5 on LIBERO, SimplerEnv, MetaWorld, and PhysicalAI-AV. GUARD achieves the best ROC-AUC on four of five unseen-task settings and ranks second on the remaining setting, improving the average unseen-task ROC-AUC by $5.73$ percentage points over the strongest competing runtime monitor while remaining within $0.19$ points of the best seen-task average. These results show that directly probing action-head dependence on multimodal evidence provides a transferable failure signal across policies, tasks, embodiments, and domains.
\end{abstract}


\section{Introduction}
\label{sec:introduction}

Large-scale robot datasets and vision-language pretraining have enabled vision-language-action (VLA) policies to follow natural-language instructions across diverse manipulation tasks and embodiments~\cite{oneill2023openx,walke2023bridgedatav2,khazatsky2024droid,brohan2022rt1,zitkovich2023rt2,kim2024openvla,octo2024,black2024pi0,shukor2025smolvla}. Recent architectures increasingly combine a pretrained vision-language backbone with diffusion- or flow-based action generation, providing an expressive mechanism for modeling continuous, multimodal robot behavior~\cite{chi2023diffusion,black2024pi0,shukor2025smolvla,romer2026flowuq}. Despite this progress, deployment remains vulnerable to novel environments, changing object appearances, compounding control errors, and task conditions that are insufficiently represented in the training data~\cite{gu2025safe,romer2025fiper,romer2026flowuq}. Under such shifts, a VLA may execute unreliable actions without an explicit indication that its prediction should not be trusted~\cite{romer2025fiper,romer2026flowuq}. Timely failure detection is therefore necessary to support intervention strategies such as stopping execution, invoking a fallback, or requesting human assistance before an error becomes unrecoverable~\cite{farid2022failure,gu2025safe,romer2025fiper}.

Failure detection is particularly challenging for generalist VLAs because an unfamiliar observation or instruction is not necessarily a failure: the policy may successfully generalize to a situation that differs from its training distribution~\cite{gu2025safe,romer2025fiper}. Task-specific detectors are also difficult to scale because generalist policies may encounter many unseen tasks for which collecting labeled success and failure rollouts is impractical~\cite{gu2025safe}. Existing monitors address parts of this problem through embedding-space novelty or likelihood scores~\cite{xu2025faildetect,gu2025safe,romer2025fiper}, temporal action consistency and progress monitoring~\cite{agia2025sentinel}, observation novelty combined with sampled action uncertainty~\cite{romer2025fiper}, or supervised probes over internal VLA features~\cite{gu2025safe}. Recent work also estimates epistemic uncertainty in flow-based VLAs through disagreement among ensembled velocity fields, but this requires maintaining multiple policy models~\cite{romer2026flowuq}. These approaches provide useful evidence about distribution shift, representation state, action variability, or model disagreement, yet they do not directly measure whether the generated action remains grounded in the particular visual and language evidence that defines the current task~\cite{gu2025safe,romer2025fiper,romer2026flowuq}.

\begin{figure}[t]
\centering
\figtarget{fig:saliency_failure_success}
\includegraphics[width=\columnwidth]{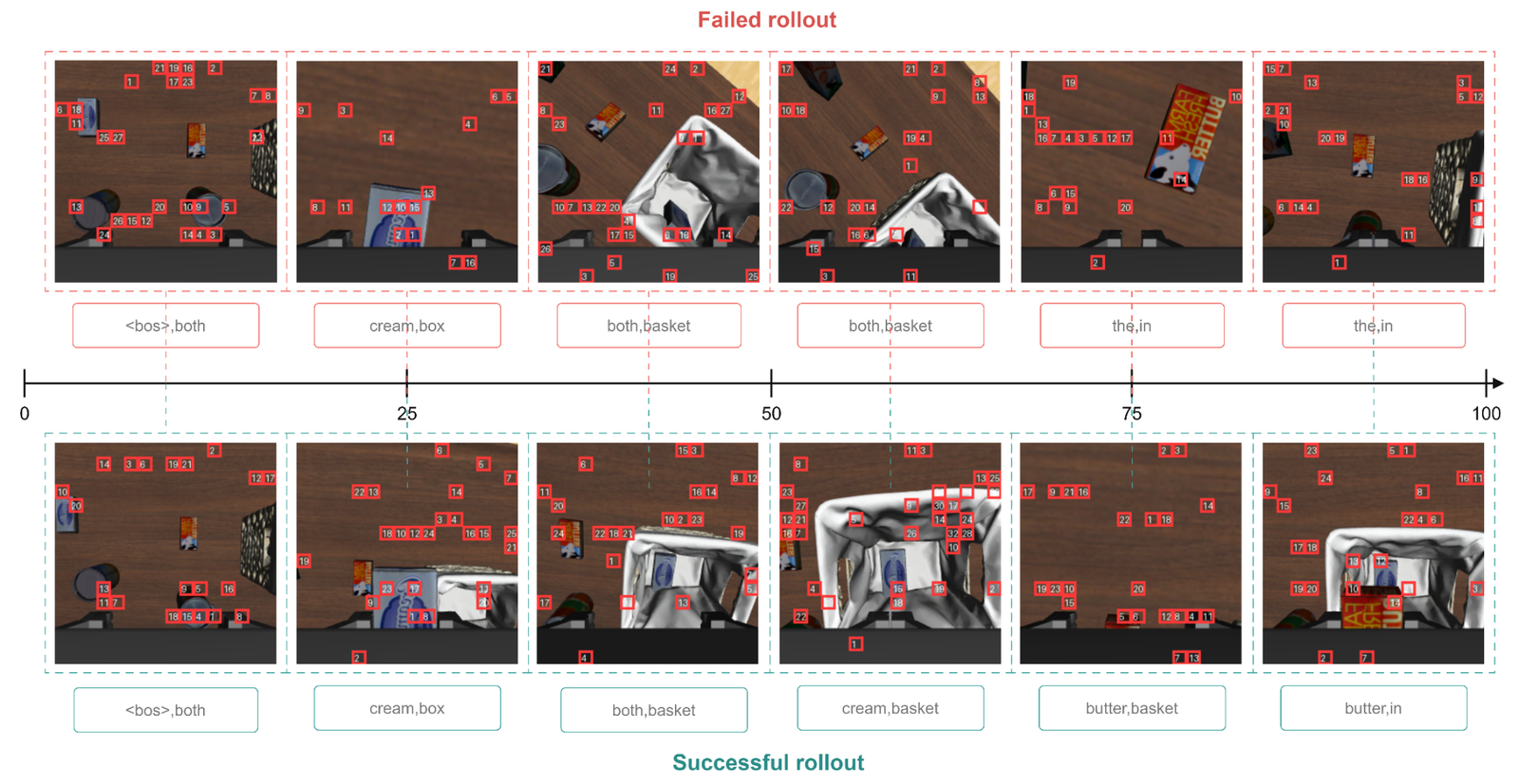}
\caption{Qualitative comparison of saliency across failed (top) and successful (bottom) rollouts for the task ``Put both the cream cheese box and the butter in the basket,'' shown at normalized progress checkpoints. Numbered boxes indicate salient visual-token locations, while the text below each frame lists the corresponding salient language tokens. In the successful rollout, saliency initially concentrates on the cream cheese box as the robot places it in the basket, and then shifts to the butter as the robot completes the second subgoal. In the failed rollout, the visual and language saliency similarly supports placing the cream cheese box in the basket, but does not subsequently focus on the butter when the second subgoal should be executed. This motivates our hypothesis that, when the conditioning entries most influential to action generation are poorly aligned with the active task evidence, ablating those entries produces only a small change in the generated action.(Please zoom in 500\%)}
\label{fig:saliency_failure_success}
\end{figure}

We study failure detection from this complementary perspective. Our central hypothesis is that a reliable action chunk should remain functionally coupled to the salient multimodal evidence specifying what the robot should perceive and do. Conversely, when the diffusion action head becomes insensitive to influential visual or language tokens, the resulting action may appear smooth or confident while being weakly grounded in the task context. The qualitative examples in \figlink{fig:saliency_failure_success} motivate this view by exposing which visual regions and language tokens influence action generation throughout successful and failed rollouts.

To operationalize this hypothesis, we introduce GUARD, a test-time failure detector for VLAs with diffusion- or flow-matching action heads. These models condition action generation on contextualized visual and language token representations, together with the robot state. GUARD intervenes on the token-aligned conditioning KV entries passed from the VLM to the action head, implemented as a KV cache used either as a layer-wise prefix  or as cross-attention in the action head, depending on the VLA architecture. Since this KV cache preserves token positions while integrating information across modalities, it supports token-level attribution and counterfactual intervention on the evidence used for action generation.

At each action-generation step, GUARD estimates which visual and language cache entries most influence the action-head response. It then constructs  \emph{counterfactual caches}, defined as controlled alternatives to the original cache in which only the most salient entries from the token positions of each modality are replaced by that modality's mean cache entries, while all remaining entries are unchanged. The action head evaluates the original cache and these visual- and language-ablated counterfactual caches from the same noisy action state. Consequently, the change in a single denoising response isolates the policy's dependence on the removed evidence.

These comparisons produce a compact diagnostic stream: \textit{sensitivity} measures the response change under ablation; \textit{attention entropy} measures how broadly influence is distributed across tokens; \textit{modality bias} compares dependence on visual and language evidence; \textit{entropy-adjusted calibration} measures deviation from an episode-specific sensitivity baseline; and \textit{grounding efficiency} relates sensitivity to attention entropy. A lightweight temporal classifier aggregates these diagnostics online. Functional conformal prediction then calibrates time-dependent alarm thresholds from classifier-score trajectories, enabling a user-controlled false-alarm tolerance~\cite{vovk2005algorithmic,lei2015conformal,angelopoulos2021gentle}. The pretrained VLA remains frozen, and the original and counterfactual cache variants are evaluated together in a batched single-step probe.

Across five task-held-out manipulation and autonomous-driving settings, GUARD ranks first on four unseen-task evaluations and second on the fifth, improving average unseen-task ROC-AUC by $5.73$ percentage points over the strongest competing state-of-the-art monitor while providing timely alarms. 


\section{Related Work}
\label{sec:relatedwork}

\paragraph{Vision-language-action policies with generative action heads.}
Recent robot policies increasingly combine pretrained vision-language representations with continuous action generation. 
RT-1 and RT-2 established large-scale vision-language-action (VLA) policies for language-conditioned control, showing that Transformer policies can benefit from diverse robot data and web-scale visual-language pretraining~\cite{brohan2022rt1,zitkovich2023rt2}. 
OpenVLA and Octo further advanced open generalist robot policies trained across multiple embodiments and manipulation tasks~\cite{kim2024openvla,octo2024}. 
A parallel trend replaces direct action regression with diffusion or flow-based action heads. 
Diffusion Policy demonstrated that iterative denoising can model multimodal continuous actions for visuomotor control~\cite{chi2023diffusion}, while Pi0 uses a flow-matching action expert conditioned on a pretrained VLM to generate action chunks for general robot control~\cite{black2024pi0}. 
SmolVLA follows this direction with an efficient VLA architecture~\cite{shukor2025smolvla}, and autonomous-driving VLAs such as Alpamayo-R1 use generative trajectory decoders for planning~\cite{wang2025alpamayo}. Related generative robot policies also study efficient VLA fine-tuning, spatially grounded representations, and diffusion-Transformer action modules~\cite{kim2025openvlaoft,qu2025spatialvla,liu2024rdt,li2024cogact}.
Our work targets this class of diffusion-action-head VLAs and asks whether failures can be detected by probing the dependence of the generated action on visual-language conditioning.

\paragraph{Runtime failure detection.}
Failure detection for learned policies is often formulated as uncertainty estimation, out-of-distribution detection, or temporal consistency monitoring~\cite{hendrycks2017baseline,lakshminarayanan2017deepensembles}. 
Representation-based methods compare test-time embeddings against training distributions using nearest-neighbor, cosine, clustering, or Mahalanobis-style scores~\cite{papernot2018deepknn,lee2018mahalanobis}. 
In robot learning, runtime monitors have also used uncertainty and conformal prediction to detect policy failures without requiring large failure datasets~\cite{farid2022failure,xu2025faildetect}. 
Recent methods for generative policies include FIPER, which combines observation novelty and action uncertainty for failure prediction~\cite{romer2025fiper}, and Sentinel/STAC, which monitors temporal action consistency and task progress during execution~\cite{agia2025sentinel}. 
SAFE studies multitask failure detection for VLAs by training lightweight probes on internal VLA features and evaluating them under task-held-out splits~\cite{gu2025safe}. 
These methods provide strong baselines, but primarily rely on embedding distance, action uncertainty, or temporal consistency.

\paragraph{Attribution and intervention.}
Gradient-based attribution estimates the local dependence of model outputs on inputs or internal representations~\cite{simonyan2013saliency,sundararajan2017integrated}. Perturbation-based explanations complement these scores by testing importance through controlled interventions on selected features~\cite{fong2017meaningful}. Studies of Transformer attention further show that raw attention weights need not provide faithful importance estimates, motivating approaches that combine attention analysis with gradients or ablations~\cite{jain2019attention,serrano2019attention,abnar2020attentionflow}.


\section{Methodology}
\label{sec:methodology}

We propose a test-time failure detection framework for vision-language-action (VLA) policies with diffusion-based action head. The key hypothesis is that a reliable action chunk should remain functionally coupled to the visual and language evidence that specifies the task. When the policy's denoising response becomes weakly dependent on the most salient multimodal tokens, the resulting action is more likely to be poorly grounded and therefore risky to execute. Our method probes this dependence without modifying the base policy. The VLA remains frozen; the detector operates by extracting a compact stream of diagnostic variables from the policy's own action-generation process.

\begin{figure}[t]
\centering
\figtarget{fig:vla_fd_architecture}
\includegraphics[width=0.98\columnwidth]{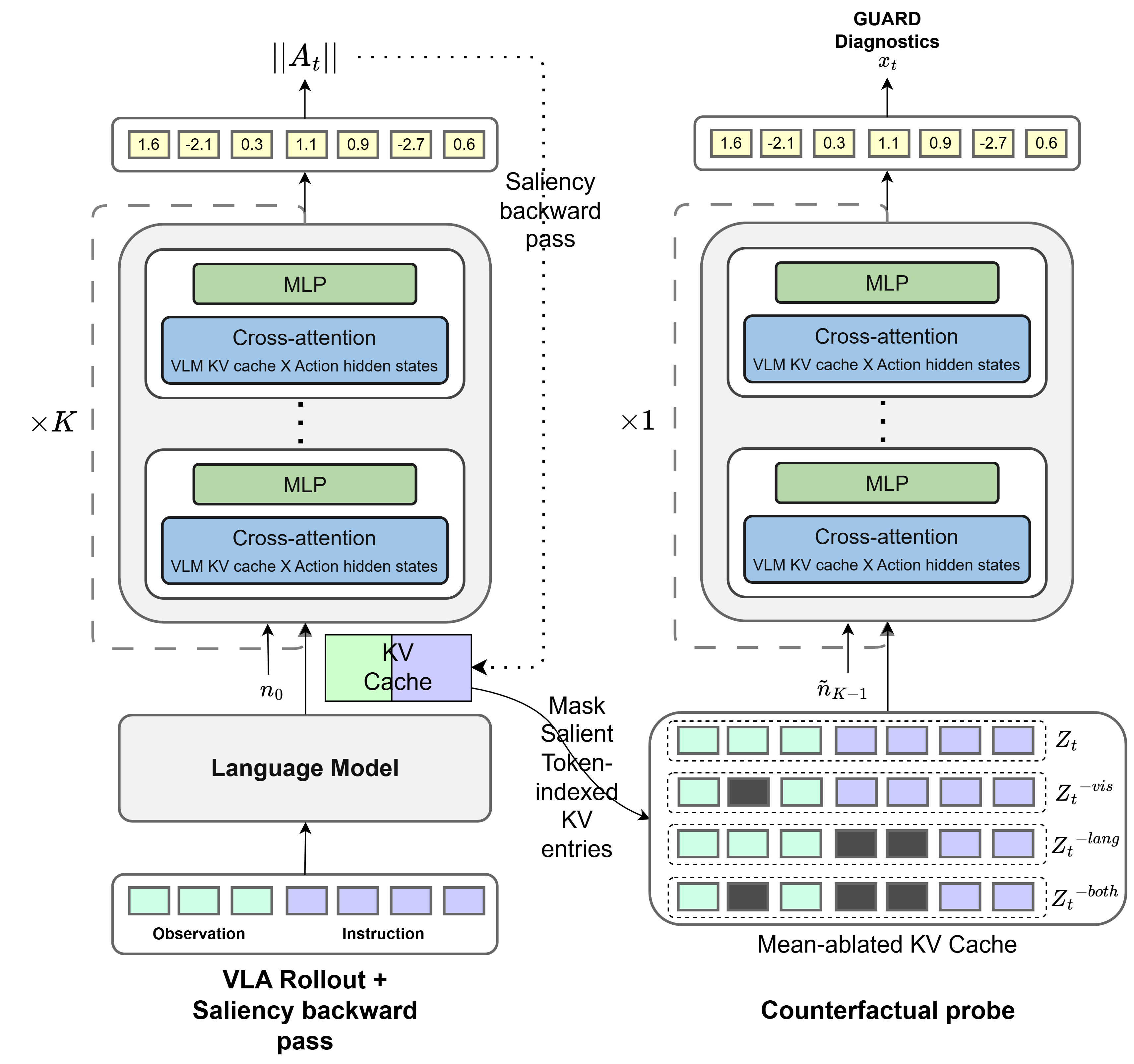}
\caption{GUARD diagnostic creation with VLA rollout, saliency backward pass, and counterfactual probe with mean ablations.}
\label{fig:vla_fd_architecture}
\end{figure}

The method has three stages. First, the frozen VLA policy performs its standard rollout to generate an action chunk. During this rollout, the VLM produces its final key--value (KV) cache, which is subsequently supplied as the conditioning context to the diffusion action head. Because the cache positions retain a one-to-one correspondence with the visual and language token positions, we estimate the influence of each token-indexed KV-cache entry on the generated action using a \textit{saliency backward pass}. Second, a lightweight counterfactual probe constructs counterfactual caches by ablating the key and value representations at the most influential $\rho_m$ fraction of cache positions within each modality with modality-specific mean key and value representations. The original and ablated final-cache variants are batched together and evaluated from the same noisy action state using a single denoising step. The resulting normal and counterfactual denoising responses are then used to compute uncertainty and grounding diagnostics. \figlink{fig:vla_fd_architecture} summarizes the full test-time pipeline for generating the diagnostics. 
Third, a lightweight temporal failure classifier consumes short windows of these diagnostics and predicts online failure risk. The classifier may use recurrent or attention-based layers, such as GRU, LSTM, or Transformer layers. 

The following subsections define these components in sequence. \subseclink{sec:surgical_saliency} defines the cache-entry influence scores used to select intervention targets. \subseclink{sec:counterfactual_probes} then defines sensitivity $S_t$ and modality bias $B_t$, while \subseclink{sec:grounding_calibration} defines attention entropy $E_t$, grounding efficiency $G_t$, the adaptive threshold $\theta_t$, calibration status $c_t$, and low-sensitivity indicator $b_t$. Finally, \subseclink{sec:safe_diagnostics} and \subseclink{sec:window_failure_classifier} combine these seven quantities into the online failure-risk prediction.


\subsection{Problem Setup}
\sectarget{sec:problem_setup}
\label{sec:problem_setup}

At control step $t$, the robot observes $o_t = \{I_t,\ell_t,s_t\}$, where $I_t$ denotes one or more camera observations, $\ell_t$ is the language instruction, and $s_t$ is the robot state when available. A pretrained VLA policy $\pi_\theta$ predicts an action chunk $A_t=[a_t,a_{t+1},\ldots,a_{t+H-1}]\in\mathbb{R}^{H\times d_a}$, where $H$ is the action horizon and $d_a$ is the action dimension~\cite{zhao2023act}. The VLM encodes the multimodal observation into its final KV cache $Z_t$, which is passed as conditioning context to the diffusion action head. We index this cache by the originating visual and language token positions:
\begin{equation}
    \begin{aligned}
    Z_t &= [z_1,z_2,\ldots,z_N]
          = f_\theta(I_t,\ell_t,s_t),\\
    z_i &= (\mathbf{k}_i,\mathbf{v}_i),
    \end{aligned}
    \label{eq:conditioning_tokens}
\end{equation}
where $z_i$ denotes the token-indexed cache entry containing the key and value representations $(\mathbf{k}_i,\mathbf{v}_i)$ associated with token position $i$.

For diffusion- or flow-style action generators, the action chunk is obtained by iterative denoising~\cite{ho2020ddpm,lipman2023flowmatching}. Let $n_0 \sim \mathcal{N}(0,I)$ and let $K$ be the number of denoising steps, with $A_t = n_K$. 
Our detector evaluates the sampled chunk, and all saliency and ablation operations are applied to the final VLM cache $Z_t$ used by the diffusion action head.

\subsection{Saliency over Token-Indexed KV-Cache}
\sectarget{sec:surgical_saliency}
\label{sec:surgical_saliency}

The first step is to identify which token-indexed entries in the final VLM KV cache locally influence the produced action chunk, following gradient-based attribution methods~\cite{simonyan2013saliency,sundararajan2017integrated}. We define the action-level scalar objective $\Phi(A_t)=\left\lVert\operatorname{vec}(A_t)\right\rVert_2$ and compute the saliency of token position $i$ as
\begin{equation}
    g_i = \left\lVert \frac{\partial \Phi(A_t)}{\partial z_i} \right\rVert_2.
    \label{eq:token_saliency}
\end{equation}
Here, $z_i=(\mathbf{k}_i,\mathbf{v}_i)$ is the token-indexed entry in the final VLM KV cache supplied to the diffusion action head. The saliency score aggregates the gradient norm over both the key and value tensors associated with position $i$. Thus, $g_i$ measures the local influence of the cache state corresponding to a specific visual or language token, while the intervention itself is performed directly in KV-cache space.

Let $\mathcal{M}_{\mathrm{vis}}$ and $\mathcal{M}_{\mathrm{lang}}$ be the visual and language token index sets. To avoid selecting only the dominant modality, we select salient positions independently within each modality:
\begin{equation}
    \begin{aligned}
    \mathcal{S}_m
    &= \operatorname{TopK}\!\left(
        \{g_i : i \in \mathcal{M}_m\},
        \left\lceil \rho_m |\mathcal{M}_m| \right\rceil
    \right),\\
    &\hspace{2em} m \in \{\mathrm{vis},\mathrm{lang}\},
    \end{aligned}
    \label{eq:topk_saliency}
\end{equation}
where $\rho_{\mathrm{vis}}$ and $\rho_{\mathrm{lang}}$ are modality-specific saliency fractions. The selected set is $\mathcal{S}=\mathcal{S}_{\mathrm{vis}}\cup\mathcal{S}_{\mathrm{lang}}$.

\subsection{Counterfactual Mean-Ablation Probes}
\sectarget{sec:counterfactual_probes}
\label{sec:counterfactual_probes}

After selecting salient positions, we test whether the action-generation dynamics actually depend on them through controlled perturbation~\cite{fong2017meaningful}. For modality $m$, define a modality mean representation
\begin{equation}
    \bar{z}_m = \frac{1}{|\mathcal{M}_m|}\sum_{i\in\mathcal{M}_m} z_i.
    \label{eq:modality_mean}
\end{equation}
We construct three ablated conditioning states: $Z_t^{-\mathrm{vis}}$, where selected visual positions are replaced by $\bar{z}_{\mathrm{vis}}$; $Z_t^{-\mathrm{lang}}$, where selected language positions are replaced by $\bar{z}_{\mathrm{lang}}$; and $Z_t^{-\mathrm{both}}$, where both selected visual and language positions are replaced. Mean replacement keeps the ablated representation on the same scale as the original feature space while removing token-specific information. The robot state is not ablated.

To make the probe comparable across original and ablated conditionings, we evaluate all responses at the same noisy action input. Given a fixed probe noise $\xi$ and probe scale $\lambda_{\mathrm{p}} \in (0,1]$, define
\begin{equation}
    \tilde{n}_{K-1}
    = (1-\lambda_{\mathrm{p}})A_t
    + \lambda_{\mathrm{p}}\xi.
    \label{eq:noisy_probe_action}
\end{equation}
A single denoising response is then computed under the original $r_t^{+}=D_\theta(\tilde{n}_{K-1},\lambda_{\mathrm{p}},Z_t)$ and ablated conditionings $r_t^{-\mathrm{vis}}=D_\theta(\tilde{n}_{K-1},\lambda_{\mathrm{p}},Z_t^{-\mathrm{vis}})$, $r_t^{-\mathrm{lang}}=D_\theta(\tilde{n}_{K-1},\lambda_{\mathrm{p}},Z_t^{-\mathrm{lang}})$, $r_t^{-\mathrm{both}}=D_\theta(\tilde{n}_{K-1},\lambda_{\mathrm{p}},Z_t^{-\mathrm{both}})$.
The sensitivity score $S_t$ is defined as the normalized change caused by ablating the selected visual-language evidence $\mathcal{S}$:
\begin{equation}
    S_t =
    \frac{\left\lVert r_t^{+} - r_t^{-\mathrm{both}} \right\rVert_2}
         {\left\lVert r_t^{+} \right\rVert_2 + \varepsilon}.
    \label{eq:sensitivity}
\end{equation}
Large $S_t$ means that the policy's denoising response changes substantially when salient evidence is ablated. Low $S_t$ means that the sampled action is weakly affected by the identified visual-language evidence, which we treat as a potential grounding failure.

We also compute a modality-bias ratio,
\begin{equation}
    B_t =
    \frac{\left\lVert r_t^{+} - r_t^{-\mathrm{vis}} \right\rVert_2}
         {\left\lVert r_t^{+} - r_t^{-\mathrm{lang}} \right\rVert_2 + \varepsilon}.
    \label{eq:bias_ratio}
\end{equation}
This ratio distinguishes failures dominated by ablation of KV cache entries in visual token position vs language token position. 

\subsection{Entropy-Normalized Grounding and Online Calibration}
\sectarget{sec:grounding_calibration}
\label{sec:grounding_calibration}

Sensitivity can be interpreted relative to the policy's attention distribution and the task-specific scale of the rollout. Because raw attention weights need not provide faithful importance estimates, GUARD uses entropy only as a complementary concentration statistic and pairs it with intervention-based sensitivity~\cite{jain2019attention,serrano2019attention,abnar2020attentionflow}. We therefore measure the entropy of the final-layer cross-attention from action queries to conditioning tokens. Let $\bar{P}_{q,i}$ be the attention probability from query $q$ to conditioning token $i$, averaged over heads. The attention entropy is
\begin{equation}
    E_t = -\frac{1}{|\mathcal{Q}|}
    \sum_{q\in\mathcal{Q}}\sum_{i=1}^{N}
    \bar{P}_{q,i}\log(\bar{P}_{q,i}+\varepsilon).
    \label{eq:attention_entropy}
\end{equation}
We define grounding efficiency as
\begin{equation}
    G_t = \frac{S_t}{E_t + \varepsilon}.
    \label{eq:grounding_efficiency}
\end{equation}
Intuitively, $G_t$ is high when the policy is sensitive to salient evidence while keeping attention relatively concentrated.

\begin{table*}[!t]
\centering
\tabtarget{tab:failure_classification_results}
\caption{Seen- and unseen-task ROC-AUC (\%) across policies and benchmarks. Bold and underlined values indicate the best and second-best result in each column, respectively.}
\label{tab:failure_classification_results}
\begingroup
\scriptsize
\setlength{\tabcolsep}{3pt}
\renewcommand{\arraystretch}{1.12}
\begin{tabular}{>{\centering\arraybackslash}m{1.85cm}|*{5}{>{\centering\arraybackslash}m{1.02cm}>{\centering\arraybackslash}m{1.02cm}|}>{\centering\arraybackslash}m{1.02cm}>{\centering\arraybackslash}m{1.02cm}}
\toprule
\rowcolor{resultheadergray}
& \multicolumn{2}{c|}{\shortstack{\textbf{Pi0}\\\textbf{Libero}}}
& \multicolumn{2}{c|}{\shortstack{\textbf{SmolVLA}\\\textbf{Libero}}}
& \multicolumn{2}{c|}{\shortstack{\textbf{Pi0}\\\textbf{SimplerEnv}}}
& \multicolumn{2}{c|}{\shortstack{\textbf{SmolVLA}\\\textbf{Meta-world}}}
& \multicolumn{2}{c|}{\shortstack{\textbf{Alpamayo}\\\textbf{Physical AI AV}}}
& \multicolumn{2}{c}{\textbf{Average}} \\
\rowcolor{resultheadergray}
\multirow{-2}{*}{\textbf{Method}}
& \textbf{Seen} & \textbf{Unseen}
& \textbf{Seen} & \textbf{Unseen}
& \textbf{Seen} & \textbf{Unseen}
& \textbf{Seen} & \textbf{Unseen}
& \textbf{Seen} & \textbf{Unseen}
& \textbf{Seen} & \textbf{Unseen} \\
\midrule
SAFE
& \underline{96.32} & 82.64
& 97.26 & 68.75
& 88.38 & 80.11
& \underline{99.83} & 85.40
& 68.47 & 64.19
& \underline{90.05} & 76.22 \\
\addlinespace[2pt]
Euclidean
& \textbf{96.65} & 74.68
& \textbf{98.61} & 69.41
& \underline{89.73} & 68.41
& 99.60 & 80.32
& 65.28 & 58.56
& 89.97 & 70.28 \\
Cosine
& 95.99 & 73.36
& \underline{98.55} & 68.35
& \textbf{90.19} & 71.32
& 99.60 & 89.59
& 66.59 & 61.27
& \textbf{90.18} & 72.78 \\
Mahalanobis
& 95.60 & 52.98
& 97.78 & 56.08
& 88.42 & 52.84
& 99.77 & 69.02
& 57.53 & 53.98
& 87.82 & 56.98 \\
PCA-kMeans
& 72.51 & 37.19
& 85.51 & 54.73
& 66.88 & 61.19
& 99.58 & 94.01
& 54.27 & 50.16
& 75.75 & 59.46 \\
\addlinespace[2pt]
logpzo
& $\sim$ & 77.38
& $\sim$ & 50.04
& 88.79 & 74.66
& 97.95 & 84.98
& 60.94 & 56.07
& 82.56 & 68.63 \\
FIPER
& $\sim$ & \underline{89.46}
& $\sim$ & \underline{84.97}
& $\sim$ & \underline{88.95}
& $\sim$ & \textbf{99.69}
& \underline{70.41} & \textbf{68.46}
& 70.41 & \underline{83.11} \\
\addlinespace[2pt]
STAC
& 66.55 & 63.91
& 64.97 & 42.35
& 60.74 & 62.21
& 92.62 & 60.34
& 51.24 & 49.38
& 67.22 & 55.64 \\
STAC-single
& 70.46 & 68.39
& 82.21 & 64.42
& 68.71 & 70.40
& 96.90 & 78.95
& 57.96 & 54.81
& 75.25 & 67.39 \\
\midrule
\rowcolor{resultrowgray}
GUARD
& 94.86 & \textbf{91.24}
& 95.41 & \textbf{91.06}
& 86.60 & \textbf{90.98}
& \textbf{99.94} & \underline{98.84}
& \textbf{73.16} & \textbf{72.08}
& 89.99 & \textbf{88.84} \\
\bottomrule
\end{tabular}
\endgroup
\end{table*}

Because the absolute scale of $S_t$ can vary across policies, tasks, and episodes, we calibrate an episode-level constant during an initial calibration window $\mathcal{C}$. For $u \in \mathcal{C}$,
\begin{equation}
    C = \frac{1}{|\mathcal{C}|}\sum_{u\in\mathcal{C}} S_u(E_u+\varepsilon)^\gamma,
    \label{eq:calibration_constant}
\end{equation}
where $\gamma$ controls the entropy correction. After calibration, the adaptive threshold is
\begin{equation}
    \theta_t = \mu\frac{C}{(E_t+\varepsilon)^\gamma},
    \label{eq:adaptive_threshold}
\end{equation}
where $\mu$ is a safety margin. $c_t = 1$ if the calibration is ongoing and $0$ otherwise. A low-sensitivity event (LSE) is recorded as $b_t=\mathbf{1}[S_t<\theta_t]$.
The LSE is therefore not an externally supplied failure label. It is an internally defined event that occurs when the current sensitivity falls below the entropy-adjusted level expected from the calibration window.

\subsection{Final GUARD Diagnostic Stream}
\sectarget{sec:safe_diagnostics}
\label{sec:safe_diagnostics}

At each action-chunk generation step, GUARD produces the seven-dimensional diagnostic vector
\begin{equation}
    x_t = [S_t,B_t,E_t,G_t,\theta_t,c_t,b_t] \in \mathbb{R}^7,
    \label{eq:seven_features}
\end{equation}
The primary diagnostics $S_t$, $E_t$, and $B_t$ characterize policy grounding and uncertainty, while the four derived parameters $\theta_t$, $b_t$, $c_t$, and $G_t$ provide the additional context used by the online temporal failure classifier.

\subsection{Window-Based Online Failure Classifier}
\sectarget{sec:window_failure_classifier}
\label{sec:window_failure_classifier}

Single-step diagnostics can be noisy because action sampling, occlusion, and transient attention shifts can affect individual probes. We therefore train a temporal classifier over short diagnostic windows. Given a rollout diagnostic sequence $x_{1:T}$, we extract fixed-length windows spaced uniformly throughout the rollout
\begin{equation}
    X_{s:s+L-1} = [x_s,x_{s+1},\ldots,x_{s+L-1}],
    \label{eq:diagnostic_window}
\end{equation}
where $L$ is the window length.

The classifier is trained with a \textit{from\_first\_lse} target. Successful episodes have negative targets at all timesteps. For a failed episode, we identify the first LSE as $t_{\mathrm{LSE}}=\min\{t:S_t<\theta_t\}$.
The timestep targets are then
\begin{equation}
    y_t =
    \begin{cases}
        0, & t < t_{\mathrm{LSE}},\\
        1, & t \geq t_{\mathrm{LSE}}.
    \end{cases}
    \label{eq:from_first_lse_targets}
\end{equation}
This avoids treating every step of a failed rollout as risky, since the failure may emerge somewhere in the middle of the rollout.

\FloatBarrier

\begin{figure*}[!t]
\centering
\figtarget{fig:online_failure_detection}
\includegraphics[width=0.98\textwidth]{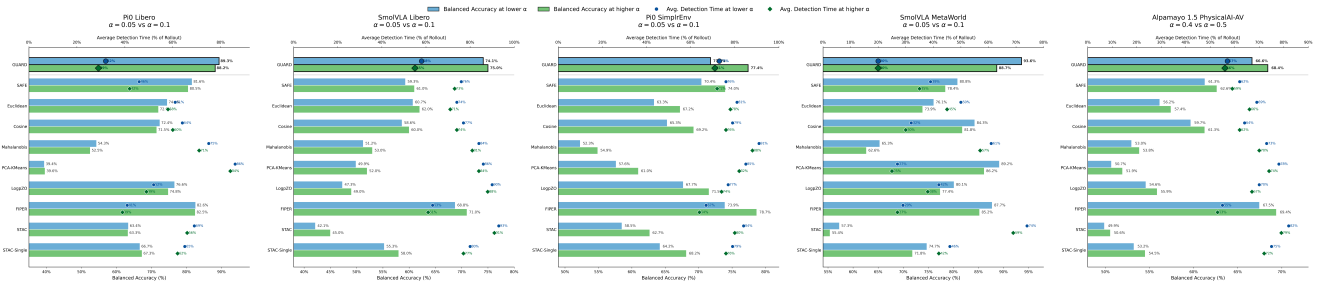}
\caption{Thresholded online failure detection across five policy--benchmark settings for unseen tasks. Blue and green bars show balanced accuracy at the lower and higher conformal levels, respectively, using the lower horizontal axis; the manipulation settings use $\alpha=0.05$ and $0.10$, while PhysicalAI-AV uses $\alpha=0.4$ and $0.5$. Blue circles and green diamonds report the corresponding average detection time as a percentage of rollout duration using the upper axis. Higher balanced accuracy and lower detection time are preferred. (Please zoom in 500\%)}
\label{fig:online_failure_detection}
\end{figure*}

\begin{table*}[!t]
\centering
\tabtarget{tab:saliency_ablation_results}
\caption{Seen- and unseen-task ROC-AUC (\%) for different cache-ablation percentages and selection modes. We keep $\rho_{\mathrm{vis}} = \rho_{\mathrm{lang}}$ and is mentioned as $\rho$ in the table. Top, random, and least denote selection of the most salient, randomly sampled, and least salient cache entries, respectively. Bold and underlined values indicate the best and second-best result in each column.}
\label{tab:saliency_ablation_results}
\begingroup
\scriptsize
\setlength{\tabcolsep}{3pt}
\renewcommand{\arraystretch}{1.12}
\begin{tabular}{>{\centering\arraybackslash}m{1.05cm}>{\centering\arraybackslash}m{1.20cm}|*{4}{>{\centering\arraybackslash}m{1.02cm}>{\centering\arraybackslash}m{1.02cm}|}>{\centering\arraybackslash}m{1.02cm}>{\centering\arraybackslash}m{1.02cm}}
\toprule
\rowcolor{resultheadergray}
\multirow{2}{*}{\shortstack{\textbf{$\rho$ (\%)}\\\textbf{(\%)}}}
& \multirow{2}{*}{\shortstack{\textbf{Selection}\\\textbf{mode}}}
& \multicolumn{2}{c|}{\shortstack{\textbf{Pi0}\\\textbf{LIBERO}}}
& \multicolumn{2}{c|}{\shortstack{\textbf{SmolVLA}\\\textbf{LIBERO}}}
& \multicolumn{2}{c|}{\shortstack{\textbf{Pi0}\\\textbf{SimplerEnv}}}
& \multicolumn{2}{c|}{\shortstack{\textbf{SmolVLA}\\\textbf{MetaWorld}}}
& \multicolumn{2}{c}{\shortstack{\textbf{Alpamayo}\\\textbf{PhysicalAI-AV}}} \\
\rowcolor{resultheadergray}
&
& \textbf{Seen} & \textbf{Unseen}
& \textbf{Seen} & \textbf{Unseen}
& \textbf{Seen} & \textbf{Unseen}
& \textbf{Seen} & \textbf{Unseen}
& \textbf{Seen} & \textbf{Unseen} \\
\midrule
\multirow{3}{*}{10}
& Top
& \textbf{94.86} & \textbf{91.24}
& \textbf{95.41} & \textbf{91.06}
& 86.60 & \underline{90.98}
& \textbf{99.94} & \underline{98.84}
& \textbf{73.16} & \textbf{72.08} \\
& Random
& 79.14 & 69.97
& 90.63 & 82.10
& 83.28 & 83.76
& \textbf{99.94} & \textbf{99.60}
& 69.17 & 68.28 \\
& Least
& 78.95 & 71.02
& 89.48 & 70.43
& 86.38 & 90.30
& 98.13 & 97.95
& 67.45 & 65.30 \\
\addlinespace[2pt]
30
& Top
& 86.62 & 77.00
& \underline{91.89} & \underline{88.28}
& \underline{86.67} & \textbf{91.04}
& 96.97 & 94.03
& \underline{72.48} & \underline{70.16} \\
\addlinespace[2pt]
50
& Top
& \underline{88.91} & \underline{78.43}
& 88.67 & 80.32
& \textbf{87.78} & 86.95
& \underline{98.93} & 92.25
& 68.71 & 67.11 \\
\bottomrule
\end{tabular}
\endgroup
\end{table*}

We train a lightweight one- or two-layer GRU, LSTM, or Transformer model on the diagnostic dataset as a binary failure-classification task using binary cross-entropy loss. At inference time, the classifier is applied online to the most recent diagnostic window. The timestep probabilities $p_t$ are converted into rollout-level scores using window mean aggregation:
\begin{align}
    q_{\mathrm{mean}} &= \frac{1}{L}\sum_{t=s}^{s+L-1}p_t,
    \label{eq:episode_scores}
\end{align}
A rollout is classified as failure-positive when the score exceeds a threshold. The threshold is determined using functional conformal prediction. Conformal prediction provides distribution-free thresholding from calibration data~\cite{vovk2005algorithmic,shafer2008tutorial}, and functional conformal methods extend this idea to curve- or trajectory-valued scores~\cite{lei2015conformal}. More details are provided in appendix.

\section{Experiments}
\label{sec:experiments}

\subsection{Evaluation Benchmarks}
\label{sec:evaluation_benchmarks}

We evaluate GUARD across four benchmarks spanning robotic manipulation and autonomous driving. LIBERO-10 evaluates Pi0 and SmolVLA on long-horizon tabletop tasks with varied scenes, objects, and instructions~\cite{liu2023libero}. SimplerEnv tests Pi0 across related manipulation tasks and embodiment settings~\cite{li2025simpler}, while MetaWorld evaluates SmolVLA on diverse object-interaction skills~\cite{yu2020metaworld}. PhysicalAI-AV extends the study to Alpamayo-1.5 under varied driving conditions~\cite{nvidia2025physicalaiav}. In every setting, complete tasks or driving-condition groups are held out for unseen-task evaluation. Additional details on rollout collection are provided in appendix.

\subsection{Evaluation Baselines}
\label{sec:evaluation_baselines}

We compare GUARD with three baseline families on identical task-held-out splits. The learned VLA-feature detector SAFE uses an LSTM to model the temporal evolution of internal VLA features~\cite{gu2025safe}. Embedding-space detectors comprise Euclidean and cosine nearest-neighbor distances~\cite{papernot2018deepknn}, class-conditional Mahalanobis distance~\cite{lee2018mahalanobis}, and PCA-KMeans cluster distance~\cite{liu2024fleet}. Runtime monitors include LogpZO for representation likelihood~\cite{xu2025faildetect}, FIPER for observation novelty and sampled-action uncertainty~\cite{romer2025fiper}, and STAC/STAC-Single for action-chunk consistency~\cite{agia2025sentinel}.

\subsection{Evaluation Metrics}
\label{sec:evaluation_metrics}

We report both threshold-free and thresholded online detection metrics. 
For overall separability, we use ROC-AUC, which measures whether failed rollouts receive higher detector scores than successful rollouts across all possible thresholds~\cite{fawcett2006roc}. 
For each rollout, we aggregate the window-level scores into a single episode score \(Q = \frac{1}{W}\sum_{w=1}^{W} q_{\mathrm{mean}}^{(w)}\), where \(q_{\mathrm{mean}}^{(w)}\) is the mean detector score in window \(w\). 
ROC-AUC is computed using these rollout-level scores and the binary success/failure labels.

For thresholded evaluation, we report balanced accuracy, defined as \(\mathrm{BalAcc}=(\mathrm{TPR}+\mathrm{TNR})/2\). 
Balanced accuracy gives equal importance to failure detection and false-alarm avoidance, which is useful when the numbers of successful and failed rollouts are imbalanced~\cite{brodersen2010balanced}. 
At inference time, a rollout is flagged when \(q_{\mathrm{mean}}\) in Eq.~\eqref{eq:episode_scores} exceeds a time-dependent threshold \(\tau_\alpha(w)\) calibrated using Functional Conformal Prediction. We also report normalized detection time \(t_{\mathrm{det}}=w^\star/W\), where \(w^\star=\min\{w:q_{\mathrm{mean}}>\tau_\alpha(w)\}\). 
Lower values indicate earlier warning within the rollout.

\section{Results}
\sectarget{sec:results}
\label{sec:results}

\subsection{Failure--Success Separability}
\label{sec:failure_success_separability}

In \tablink{tab:failure_classification_results}, we report ROC-AUC based on the episode score \(Q\). The embedding-space detectors are strong on seen tasks: Euclidean and cosine distance lead three manipulation settings and attain the best overall seen-task result. Their performance, however, drops substantially on held-out tasks. This indicates that the geometry of the pretrained feature space is informative about failures within the training-task distribution, but is less reliable when task semantics change.

The learned SAFE baseline improves robustness over most fixed-distance scores on unseen tasks, suggesting that temporal evolution in the internal VLA features provides useful failure information. Among the runtime monitors, FIPER is the strongest competitor and achieves the best unseen result on MetaWorld with SmolVLA. In contrast, STAC and STAC-Single are less consistent across settings, showing that action-chunk instability alone does not provide a sufficiently general failure signal.

GUARD achieves the strongest and most consistent task generalization. It ranks first on four of the five unseen settings and second on the remaining MetaWorld setting, yielding an unseen-task average of $88.84\%$---$5.73$ percentage points above FIPER and $12.62$ points above SAFE. Importantly, this improvement does not sacrifice seen-task performance: GUARD remains within $0.19$ points of the best seen-task average. Its seen--unseen gap is also an order of magnitude smaller than those of the learned-feature and nearest-neighbor baselines.

The gains are consistent across two VLA architectures on Libero, across embodiments in SimplerEnv and MetaWorld, and in the substantially different PhysicalAI-AV driving domain. We attribute this robustness to GUARD's direct measurement of whether the action head remains dependent on salient multimodal evidence. Whereas embedding detectors model where a rollout lies in feature space and consistency monitors measure how actions vary, GUARD probes the action-generation mechanism itself through counterfactual cache ablation.

\subsection{Online Failure Detection}
\label{sec:online_failure_detection}

\figlink{fig:online_failure_detection} evaluates thresholded online detection using balanced accuracy and average detection time. Balanced accuracy measures the quality of the final failure decision while accounting equally for successful and failed rollouts, whereas detection time measures when the first alarm is raised as a percentage of rollout duration. A useful runtime monitor must perform well on both axes.

GUARD provides the strongest balanced-accuracy performance in three of the five settings. It leads all baselines for both operating points on Pi0 Libero, SmolVLA Libero, and SmolVLA MetaWorld, demonstrating that the ROC-AUC gains in \tablink{tab:failure_classification_results} translate into reliable thresholded decisions. The advantage is especially clear on Pi0 Libero and MetaWorld, where GUARD separates itself from both learned feature probes and runtime monitors. On Pi0 SimplerEnv and Alpamayo PhysicalAI-AV, FIPER attains the highest balanced accuracy, but GUARD remains close to the best result. Thus, GUARD is consistently competitive even when the policy architecture, embodiment, and domain change.

The detection-time results show a similar pattern. GUARD raises the earliest alarms on Pi0 Libero, SmolVLA Libero, and MetaWorld, detecting failures after roughly one-fifth to three-fifths of the rollout depending on the setting. On Pi0 Libero, it warns about ten percentage points of rollout progress earlier than FIPER while also achieving higher balanced accuracy. On MetaWorld, GUARD detects at approximately $20\%$ progress, ahead of the next-fastest competitive methods, while retaining the best accuracy. FIPER is slightly earlier on SimplerEnv and PhysicalAI-AV, but the difference is small in the driving setting, where GUARD remains the closest accuracy competitor.

\subsection{Importance of Saliency-based Ablation}
\label{sec:importance_of_saliency_based_ablation}

\tablink{tab:saliency_ablation_results} isolates the effects of both cache-entry selection and ablation percentage. The default configuration, which ablates the top $10\%$ most salient entries within each modality, achieves the highest average performance on both seen and unseen tasks. This result indicates that GUARD benefits from a focused counterfactual intervention rather than from perturbing the conditioning cache indiscriminately.

At a fixed $10\%$ ablation rate, selecting the most salient entries substantially outperforms random and least-salient selection. Averaged across settings, top-saliency ablation improves unseen-task ROC-AUC by $8.10$ percentage points over random selection and $9.84$ points over least-salient selection. It exceeds least-salient ablation in every seen and unseen task and surpasses random ablation eight of ten times, while tying one. The only exception is MetaWorld unseen-task evaluation, where random selection is slightly higher. The overall pattern confirms that the detector relies on identifying action-relevant evidence, rather than merely benefiting from noise injected into the cache.

The ablation percentage also matters. Increasing top-saliency ablation from $10\%$ to $30\%$ reduces the average seen- and unseen-task results, and increasing it to $50\%$ causes a larger decline in unseen-task performance. Four of the five policy--benchmark settings favor the $10\%$ configuration. SimplerEnv is the main exception: $30\%$ gives a marginally higher unseen result, while $50\%$ gives the highest seen result but generalizes less effectively. Thus, ablating more salient entries does not consistently strengthen the diagnostic signal.

These findings support the design of GUARD. Ablating too few unimportant entries produces little change in the denoising response, whereas random selection mixes relevant and irrelevant evidence and yields an inconsistent signal. Conversely, ablating a large fraction of the cache can remove excessive task context, making the response reflect broad conditioning corruption rather than dependence on a compact set of salient tokens. The top-$10\%$ configuration provides the most reliable balance.

\subsection{Efficiency and Deployability}
\label{sec:efficiency_and_deployability}

GUARD incurs only modest computational overhead: $1.14\times$ for Pi0 and $1.27\times$ for SmolVLA. The added cost comes from one saliency backward pass and one batched single-step counterfactual probe, while the lightweight temporal classifier is negligible. Overall, GUARD does not introduce prohibitive overhead and remains suitable for deployment; a detailed stage-wise analysis is provided in the appendix.

\section{Conclusion}
\label{sec:limitation_and_conclusion}

GUARD reframes VLA failure detection as a question of functional grounding rather than merely input novelty, representation distance, or action variability. Our results indicate that testing how action generation responds when its most influential visual and language evidence is removed provides a failure signal that remains informative across tasks, policies, embodiments, and domains. Temporal aggregation and conformal prediction turn local grounding changes into actionable online warnings.

\paragraph{Limitations.}
GUARD is currently applicable only for VLAs with diffusion-based action head, and therefore does not directly apply to autoregressive VLAs. Extending it to autoregressive action generation is a direction for future work. In addition, the saliency backward pass requires retaining intermediate activations of the action head and increases peak VRAM relative to standard policy inference; memory-efficient gradient computation and selective probing could reduce this cost.

\bibliography{aaai2027}


\onecolumn
\appendix
\raggedbottom
\section*{Appendix}
\label{app:appendix}

\section{Mathematical Intuition and Properties of GUARD Diagnostics}
\label{app:safe_diagnostic_properties}

This section gives mathematical intuition for the GUARD saliency diagnostics defined in the methodology. These results should not be read as a proof that the detector must perfectly separate success and failure. Such a theorem would require strong assumptions about the robot, environment, task distribution, and VLA policy. Instead, the analysis characterizes how the diagnostics reflect the mechanisms encoded in their definitions: sensitivity captures counterfactual dependence on salient evidence, entropy describes attention concentration, grounding efficiency combines these two effects, and the adaptive threshold compares sensitivity against an episode-calibrated operating regime.

\subsection{Local Influence Interpreted as Gradient Saliency}
\label{app:gradient_saliency}

Recall the action-level objective
\begin{equation}
    \Phi(A_t) = \|\operatorname{vec}(A_t)\|_2,
    \label{eq:app_action_objective}
\end{equation}
and the KV-cache-entry saliency score
\begin{equation}
    g_i = \left\|\frac{\partial \Phi(A_t)}{\partial z_i}\right\|_2 .
    \label{eq:app_cache_entry_saliency}
\end{equation}
The purpose of Eq.~\eqref{eq:app_cache_entry_saliency} is to estimate how strongly the final action chunk changes, to first order, when an entry at a visual or language position in the final KV cache is perturbed.

\paragraph{Property 1: first-order influence bound.}
Assume that the mapping from the final conditioning KV cache to the action objective is differentiable in a neighborhood of $Z_t$. Let $\delta z_i$ be a small perturbation applied only to the cache entry at position $i$, with $\|\delta z_i\|_2 \leq \eta$. Then
\begin{equation}
    \left|\Phi(A_t(z_i+\delta z_i)) - \Phi(A_t(z_i))\right|
    \leq \eta g_i + O(\eta^2).
    \label{eq:app_saliency_bound}
\end{equation}
\emph{Proof.} By Taylor expansion,
\begin{equation}
    \Phi(A_t(z_i+\delta z_i))
    = \Phi(A_t(z_i))
    + \left\langle \frac{\partial \Phi(A_t)}{\partial z_i},\delta z_i\right\rangle
    + O(\|\delta z_i\|_2^2).
\end{equation}
Taking absolute values and applying Cauchy--Schwarz gives
\begin{equation}
    \left|\left\langle \frac{\partial \Phi(A_t)}{\partial z_i},\delta z_i\right\rangle\right|
    \leq
    \left\|\frac{\partial \Phi(A_t)}{\partial z_i}\right\|_2
    \|\delta z_i\|_2
    \leq \eta g_i.
\end{equation}
This proves Eq.~\eqref{eq:app_saliency_bound}. \hfill$\square$

This result justifies using $g_i$ as a local influence score. A large $g_i$ means that, for the same perturbation budget, the KV-cache entry at position $i$ can produce a larger first-order change in the action objective than an entry with smaller $g_i$.

\subsection{Why Use Modality-Balanced Top-$K$ Selection?}
\label{app:topk_selection}

The method selects salient entries separately within the sets of visual and language positions in the final KV cache:
\begin{equation}
    \mathcal{S}_m = \operatorname{TopK}\!\left(
        \{g_i:i\in\mathcal{M}_m\},
        \left\lceil \rho_m |\mathcal{M}_m| \right\rceil
    \right),\qquad
    m\in\{\mathrm{vis},\mathrm{lang}\}.
    \label{eq:app_topk}
\end{equation}

\paragraph{Property 2: optimality under per-modality cardinality constraints.}
Let $k_m=\lceil\rho_m|\mathcal{M}_m|\rceil$. Among all subsets $\mathcal{T}_m\subseteq\mathcal{M}_m$ of size $k_m$, Eq.~\eqref{eq:app_topk} maximizes the total first-order saliency mass
\begin{equation}
    \sum_{i\in\mathcal{T}_m} g_i.
\end{equation}
\emph{Proof.} Suppose a selected set $\mathcal{T}_m$ of size $k_m$ is not the top-$k_m$ set. Then there exist $i\notin\mathcal{T}_m$ and $j\in\mathcal{T}_m$ such that $g_i>g_j$. Replacing $j$ with $i$ strictly increases the sum. Repeating this exchange yields the top-$k_m$ set. \hfill$\square$

The modality-balanced constraint is important because the numbers of visual and language cache positions and their gradient scales may differ. A joint top-$K$ selection could select almost all entries from one modality, whereas Eq.~\eqref{eq:app_topk} guarantees that KV-cache entries at both visual and language positions are probed.

\subsection{Mean Ablation as Minimal-Distortion Information Removal}
\label{app:mean_ablation}

For each modality $m$, the ablation value is the modality mean
\begin{equation}
    \bar{z}_m = \frac{1}{|\mathcal{M}_m|}\sum_{i\in\mathcal{M}_m}z_i.
    \label{eq:app_modality_mean}
\end{equation}
Mean replacement removes position-specific information from the selected KV-cache entries while keeping the replacement on the scale of the original modality representation.

\begin{figure}[!t]
    \centering
    \figtarget{fig:lm_expert_cross_attention_layers}
    \begin{minipage}[t]{0.49\textwidth}
        \centering
        \includegraphics[width=\linewidth]{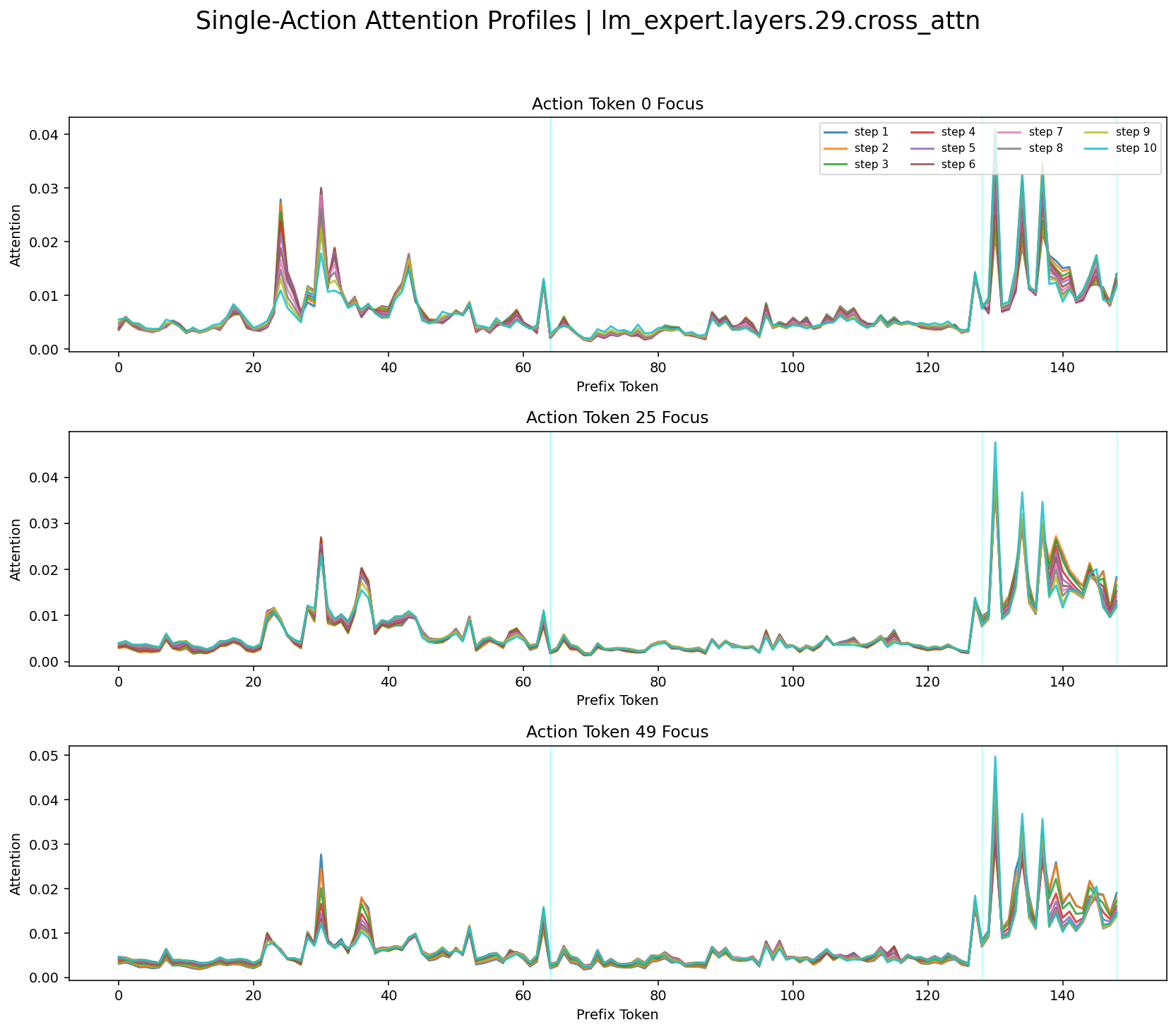}
        \textbf{(a)} Action head layer 29
    \end{minipage}\hfill
    \begin{minipage}[t]{0.49\textwidth}
        \centering
        \includegraphics[width=\linewidth]{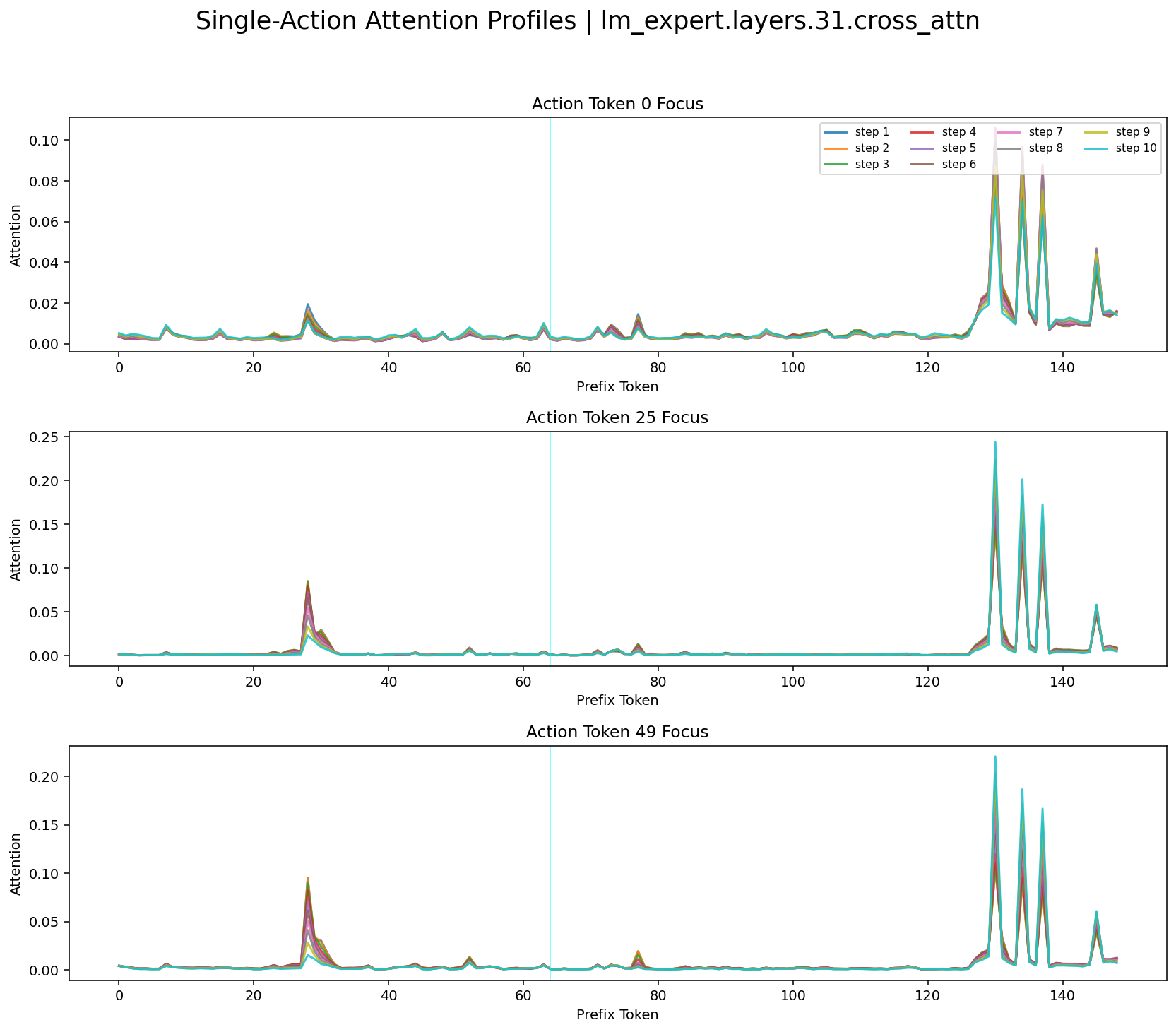}
        \textbf{(b)} Action head layer 31
    \end{minipage}
    \caption{Single-action cross-attention profiles over conditioning positions for action-query positions 0, 25, and 49 across ten denoising steps of SmolVLA on LIBERO spatial task. Panels (a) and (b) show the profiles from expert layers 29 and 31, respectively. The blue vertical lines delineate KV-cache positions associated with image input 1, image input 2, and the language instruction, in that order.}
    \label{fig:lm_expert_cross_attention_layers}
\end{figure}

The cross-attention profiles in \figlink{fig:lm_expert_cross_attention_layers} show how strongly each action query attends to the KV-cache entries aligned with visual and language positions. The attention magnitudes exhibit distinct modality-dependent patterns across these positions, indicating that visual and language entries need not share a common representation scale or attention regime. We therefore compute the replacement mean independently within each modality, preserving its characteristic scale while removing position-specific information from the selected entries.

\paragraph{Property 3: least-squares constant representative.}
Among all constant replacement vectors $c\in\mathbb{R}^d$, the mean vector in Eq.~\eqref{eq:app_modality_mean} minimizes the total squared reconstruction error
\begin{equation}
    J(c)=\sum_{i\in\mathcal{M}_m}\|z_i-c\|_2^2.
    \label{eq:app_mean_objective}
\end{equation}
\emph{Proof.} Differentiating $J(c)$ gives
\begin{equation}
    \nabla_c J(c)=2\sum_{i\in\mathcal{M}_m}(c-z_i).
\end{equation}
Setting the gradient to zero yields
\begin{equation}
    |\mathcal{M}_m|c = \sum_{i\in\mathcal{M}_m}z_i,
\end{equation}
so $c=\bar{z}_m$. Since $J(c)$ is convex in $c$, this stationary point is the global minimizer. \hfill$\square$

Thus, mean ablation is a conservative counterfactual: it removes the identity-specific information carried by the selected KV-cache entries while minimizing the magnitude of the replacement shift within the modality.

\subsection{Counterfactual Denoising Sensitivity}
\label{app:sensitivity}

The probe evaluates the original and ablated conditionings at the same noisy action input
\begin{equation}
    \tilde{n}_{K-1}
    =(1-\lambda_{\mathrm{p}})A_t
    +\lambda_{\mathrm{p}}\xi,
    \label{eq:app_noisy_probe}
\end{equation}
and computes responses $r_t^{+}$ and $r_t^{-\mathrm{both}}$. The sensitivity diagnostic is
\begin{equation}
    S_t=
    \frac{\|r_t^{+}-r_t^{-\mathrm{both}}\|_2}
         {\|r_t^{+}\|_2+\varepsilon}.
    \label{eq:app_sensitivity}
\end{equation}

\paragraph{Property 4: sensitivity estimates local dependence on ablated evidence.}
Let $D_\theta(\tilde{n}_{K-1},\lambda_{\mathrm{p}},Z)$ be differentiable with respect to $Z$. Let $\Delta Z=Z_t^{-\mathrm{both}}-Z_t$. A first-order expansion gives
\begin{equation}
    r_t^{-\mathrm{both}}
    = r_t^{+}+J_Z\Delta Z+O(\|\Delta Z\|_2^2),
    \label{eq:app_response_taylor}
\end{equation}
where $J_Z$ is the Jacobian of the denoising response with respect to the conditioning state. Therefore,
\begin{equation}
    S_t
    =
    \frac{\|J_Z\Delta Z\|_2}{\|r_t^{+}\|_2+\varepsilon}
    +O(\|\Delta Z\|_2^2).
    \label{eq:app_sensitivity_jacobian}
\end{equation}

This explains why $S_t$ is a counterfactual grounding diagnostic. If the denoising response strongly depends on the selected visual-language evidence, then ablating those positions produces a large response change. If the response barely changes, the sampled action is locally insensitive to the evidence that the saliency step identified as important.

\paragraph{Property 5: scale normalization.}
Ignoring the small stabilizer $\varepsilon$, $S_t$ is invariant to a common positive scaling of denoising responses. If $r_t^{+}$ and $r_t^{-\mathrm{both}}$ are both multiplied by $a>0$, then
\begin{equation}
    \frac{\|ar_t^{+}-ar_t^{-\mathrm{both}}\|_2}{\|ar_t^{+}\|_2}
    =
    \frac{\|r_t^{+}-r_t^{-\mathrm{both}}\|_2}{\|r_t^{+}\|_2}.
\end{equation}
This normalization makes sensitivity less dependent on the absolute magnitude of the denoising vector.

\subsection{Modality-Bias Ratio}
\label{app:bias_ratio}

The modality-bias diagnostic is
\begin{equation}
    B_t=
    \frac{\|r_t^{+}-r_t^{-\mathrm{vis}}\|_2}
         {\|r_t^{+}-r_t^{-\mathrm{lang}}\|_2+\varepsilon}.
    \label{eq:app_bias}
\end{equation}
This ratio is not a failure score by itself. Instead, it describes whether the denoising response changes more under visual ablation or language ablation.

When $B_t>1$, visual ablation changes the response more than language ablation. When $B_t<1$, language ablation has the larger effect. Since manipulation and driving failures can arise from different sources---visual ambiguity, language mismatch, or multimodal disagreement---$B_t$ gives the temporal classifier information about the source of instability rather than only its magnitude.

\subsection{Attention Entropy and Grounding Efficiency}
\label{app:entropy_grounding}

The attention entropy diagnostic is
\begin{equation}
    E_t=-\frac{1}{|\mathcal{Q}|}
    \sum_{q\in\mathcal{Q}}\sum_{i=1}^{N}
    \bar{P}_{q,i}\log(\bar{P}_{q,i}+\varepsilon).
    \label{eq:app_entropy}
\end{equation}
For each query $q$, the vector $\bar{P}_{q,1:N}$ is a probability distribution over conditioning positions in the final KV cache.

\paragraph{Property 6: entropy range.}
Ignoring $\varepsilon$ for notation, for each query $q$,
\begin{equation}
    0\leq -\sum_{i=1}^{N}\bar{P}_{q,i}\log \bar{P}_{q,i}\leq \log N.
\end{equation}
Consequently,
\begin{equation}
    0\leq E_t\leq \log N.
\end{equation}
\emph{Proof.} Shannon entropy is nonnegative and is maximized by the uniform distribution. For the uniform distribution, $\bar{P}_{q,i}=1/N$, so the entropy is $\log N$. Averaging over queries preserves the same bounds. \hfill$\square$

The grounding-efficiency diagnostic is
\begin{equation}
    G_t=\frac{S_t}{E_t+\varepsilon}.
    \label{eq:app_grounding_efficiency}
\end{equation}
It is monotone increasing in sensitivity and monotone decreasing in entropy:
\begin{equation}
    \frac{\partial G_t}{\partial S_t}=\frac{1}{E_t+\varepsilon}>0,
    \qquad
    \frac{\partial G_t}{\partial E_t}=-\frac{S_t}{(E_t+\varepsilon)^2}\leq 0.
    \label{eq:app_grounding_monotonicity}
\end{equation}
Thus, for fixed sensitivity, diffuse attention reduces grounding efficiency; for fixed entropy, stronger counterfactual dependence increases grounding efficiency.

\begin{figure}[t]
    \centering
    \figtarget{fig:app_entropy_sensitivity_phase}
    \includegraphics[width=0.96\textwidth]{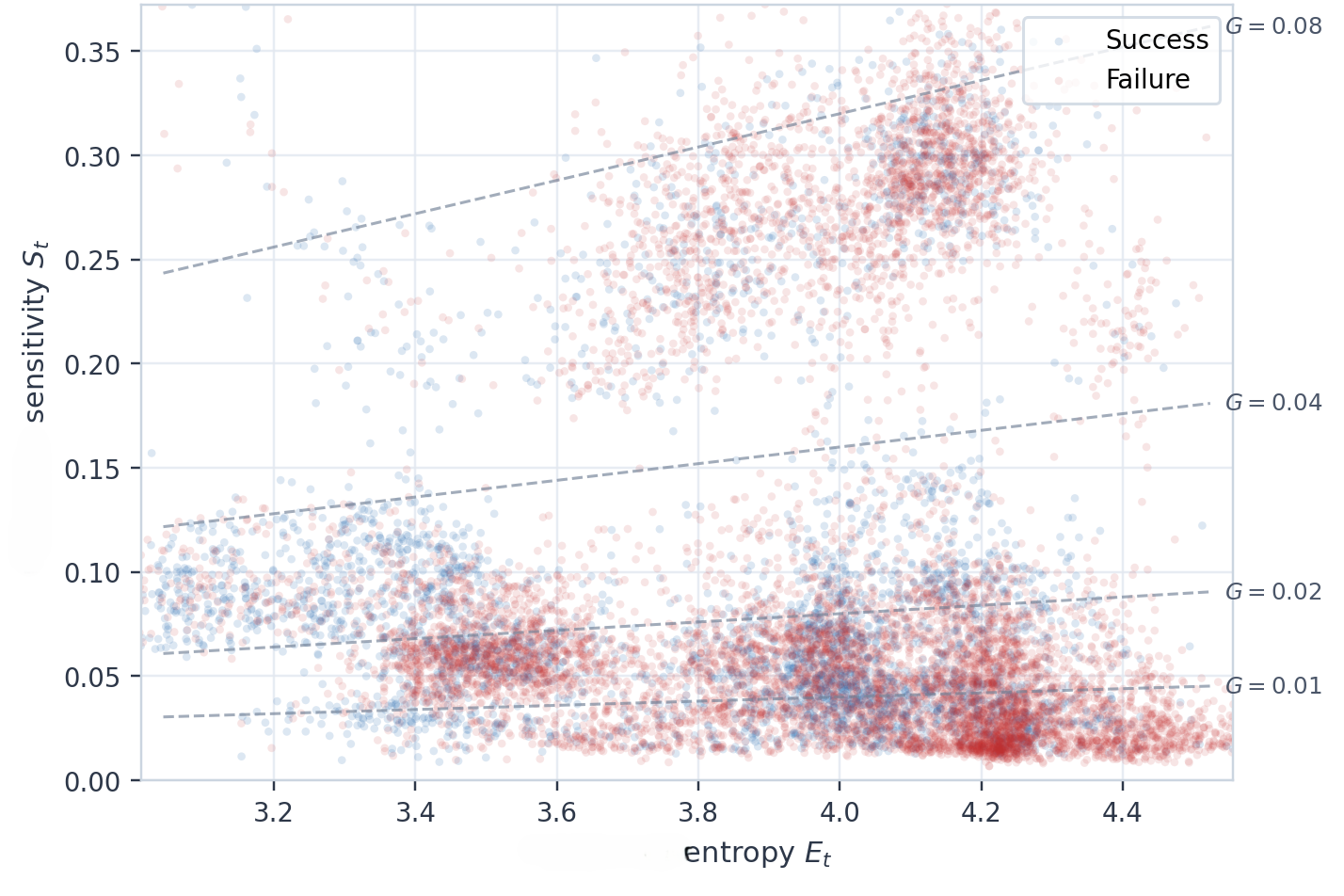}
    \caption{Entropy--sensitivity phase plot for surgical saliency diagnostics. Each point corresponds to one action-chunk generation step. The horizontal axis shows attention entropy $E_t$, while the vertical axis shows surgical sensitivity $S_t$. Dashed curves denote iso-contours of grounding efficiency $G_t = S_t/(E_t+\varepsilon)$. The plot illustrates the operating regimes induced by the diagnostic definition: high sensitivity with low entropy corresponds to high grounding efficiency, while low sensitivity with high entropy corresponds to weak grounding efficiency.}
    \label{fig:app_entropy_sensitivity_phase}
\end{figure}

\figlink{fig:app_entropy_sensitivity_phase} gives a geometric interpretation of Eq.~\eqref{eq:app_grounding_efficiency}. The upper-left region corresponds to strong grounding: the denoising response changes substantially when salient evidence is removed, while attention remains concentrated. The lower-right region corresponds to weak grounding: the response is insensitive to salient evidence and attention is diffuse. The dashed iso-contours show points with equal grounding efficiency. Moving upward increases $G_t$ because the policy response becomes more sensitive to salient evidence; moving rightward decreases $G_t$ because the same sensitivity is distributed over a more diffuse attention pattern.

This visualization reinforces the interpretation of $G_t$ as an instantaneous grounding-efficiency diagnostic and clarifies the advantage of interpreting sensitivity and entropy together. A high value of $S_t$ is most meaningful when attention entropy is low or moderate; conversely, a moderate value of $S_t$ may be less reliable when the attention distribution is highly diffuse. Thus, GUARD saliency does not merely measure whether the policy changes under ablation, but whether that change occurs under a concentrated multimodal attention regime.

\subsection{Online Calibration and the Low-Sensitivity Event}
\label{app:calibration}

The calibration equations are
\begin{equation}
    C=\frac{1}{|\mathcal{C}|}\sum_{u\in\mathcal{C}}S_u(E_u+\varepsilon)^\gamma,
    \label{eq:app_calibration_constant}
\end{equation}
\begin{equation}
    \theta_t=\mu\frac{C}{(E_t+\varepsilon)^\gamma},
    \label{eq:app_threshold}
\end{equation}
and
\begin{equation}
    b_t=\mathbf{1}[S_t<\theta_t].
    \label{eq:app_lse}
\end{equation}

\paragraph{Property 7: equivalent normalized-threshold test.}
Eq.~\eqref{eq:app_lse} is equivalent to
\begin{equation}
    \frac{S_t(E_t+\varepsilon)^\gamma}{C}<\mu.
    \label{eq:app_normalized_lse}
\end{equation}
\emph{Proof.} Since $E_t+\varepsilon>0$, multiply both sides of $S_t<\mu C/(E_t+\varepsilon)^\gamma$ by $(E_t+\varepsilon)^\gamma/C$. \hfill$\square$

This shows that the adaptive threshold compares the current entropy-adjusted sensitivity to its calibration-window baseline. The constant $C$ absorbs episode- and task-specific scale, while $\mu$ controls the safety margin.

\begin{figure}[t]
\centering
\figtarget{fig:failed_episodes_first_lse_aligned}
\includegraphics[width=0.90\textwidth]{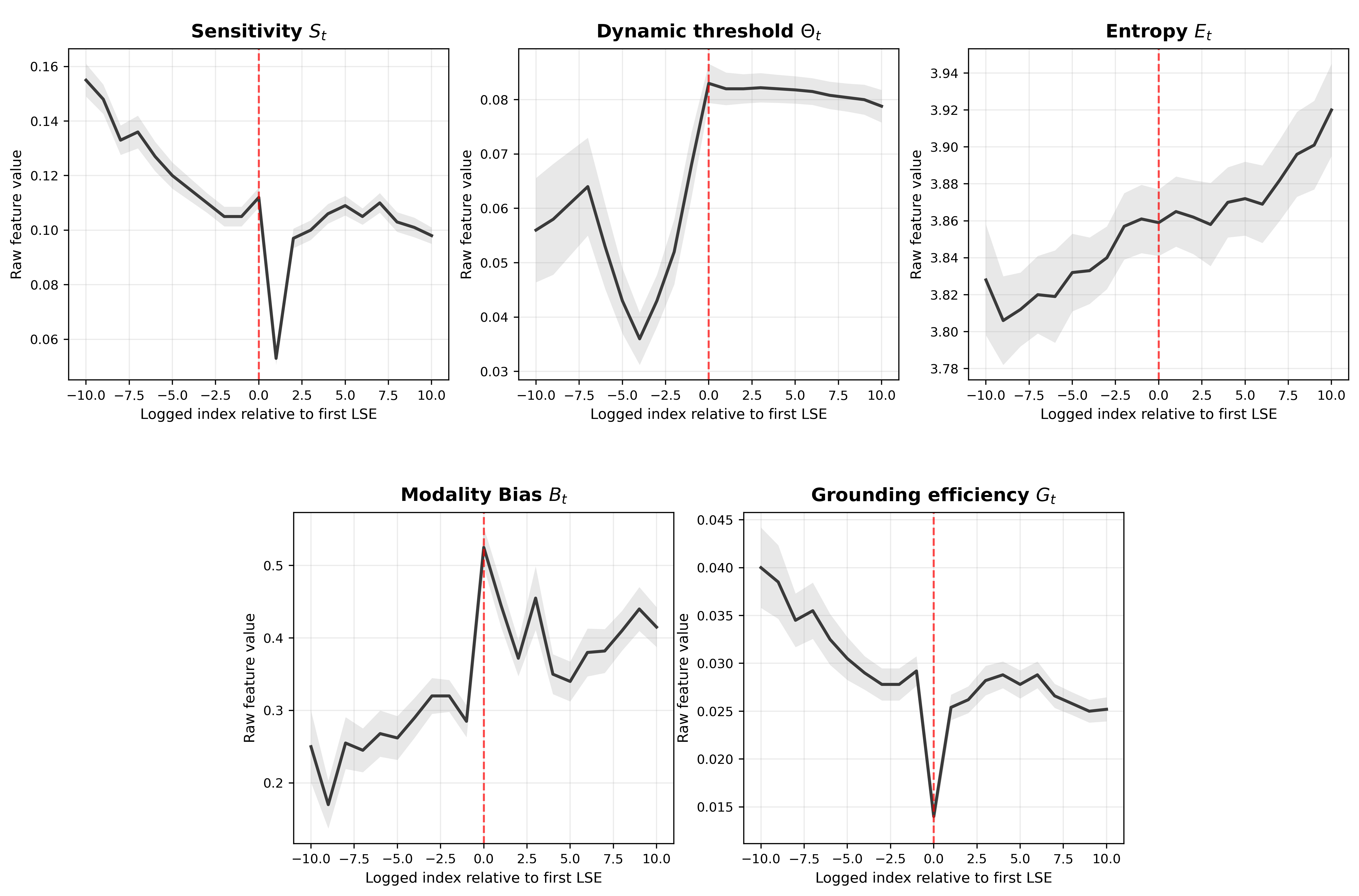}
\caption{Diagnostic trajectories for failed Pi0--LIBERO rollouts aligned at their first low-sensitivity event (LSE). Relative index zero, marked by the red dashed line, denotes the first LSE, and the shaded regions show variability across aligned episodes. Sensitivity $S_t$ decreases before the event, drops sharply immediately afterward, and only partially recovers at a lower level. The dynamic threshold $\theta_t$ rises abruptly at the event and remains elevated, with a gradual decline later in the rollout. Attention entropy $E_t$ follows an overall increasing trend, indicating progressively more diffuse attention after the first LSE. Modality bias $B_t$ spikes at the event and remains higher afterward despite short-term fluctuations, indicating a stronger imbalance between visual and language influence. Grounding efficiency $G_t$ collapses at the event and recovers only partially, remaining below its early-rollout level. Together, these trends show that the first LSE coincides with a broader shift toward weaker, more diffuse, and less balanced grounding rather than an isolated threshold crossing.}
\label{fig:failed_episodes_first_lse_aligned}
\end{figure}

\FloatBarrier

\section{Functional Conformal Prediction for Online Failure Detection}
\sectarget{app:functional_conformal_prediction}
\label{app:functional_conformal_prediction}

For rollout $i$, let $x_{i,0:T_i}$ denote the seven-dimensional GUARD diagnostic sequence from Eq.~\eqref{eq:seven_features}, and let $y_i\in\{0,1\}$ denote the trajectory-level success/failure label. The online detector has two stages: a temporal classifier converts the diagnostic history into a causal failure-score curve, and functional conformal prediction calibrates a time-varying upper threshold from successful seen-task calibration rollouts~\cite{vovk2005algorithmic,lei2015conformal,angelopoulos2021gentle}. The objective is to detect failed rollouts early while limiting alarms on successful rollouts.

\begin{figure}[!t]
\centering
\figtarget{fig:vla_fd_classifier}
\includegraphics[width=0.45\textwidth]{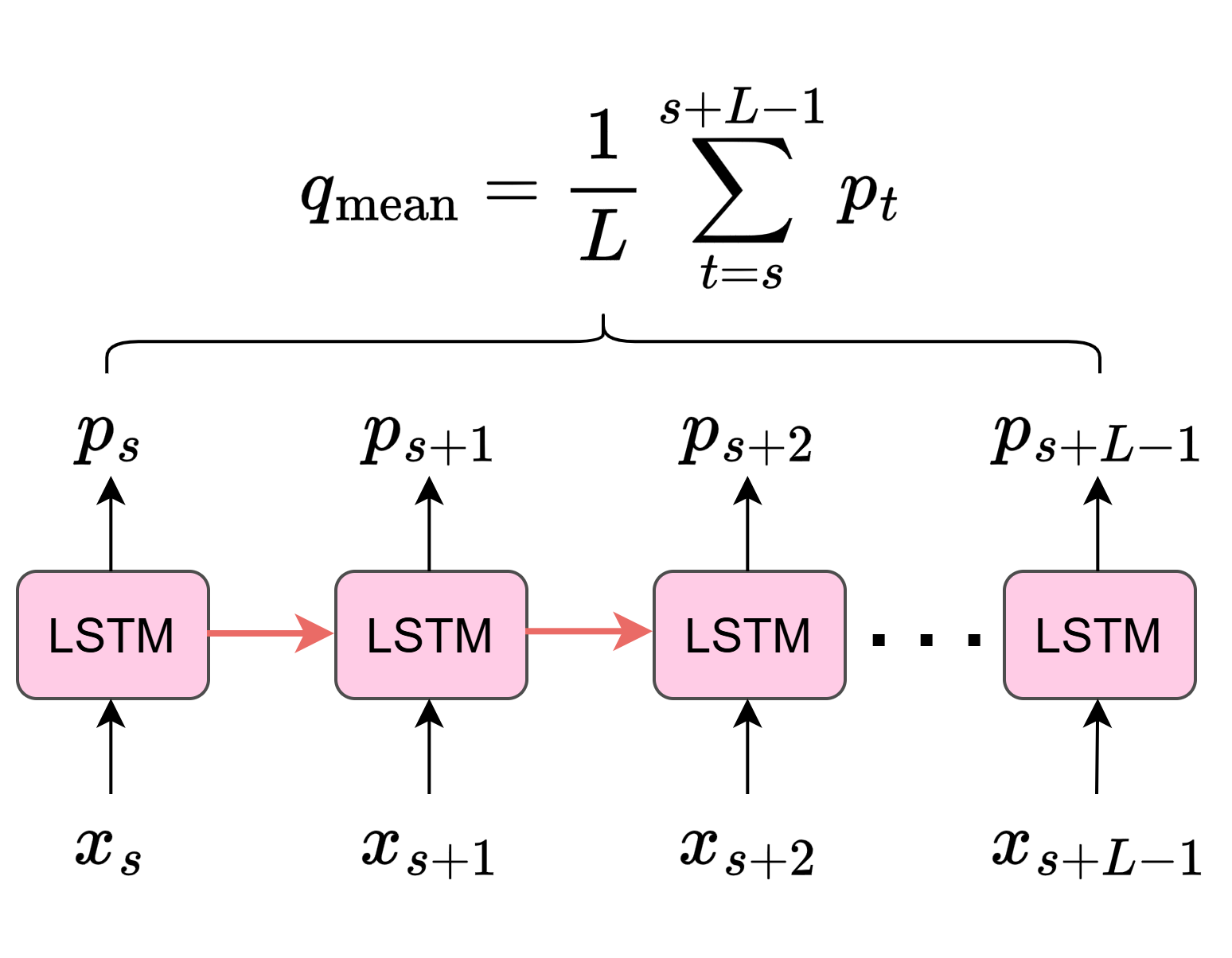}
\caption{Online failure-classifier module. At each timestep, the most recent causal window of normalized GUARD diagnostics is processed by the temporal classifier and aggregated into a scalar failure score. Dense causal evaluation produces the score trajectory subsequently calibrated by functional conformal prediction.}
\label{fig:vla_fd_classifier}
\end{figure}

\paragraph{Dense causal score curve.}
Although training uses sampled fixed-length windows, online timing is evaluated densely and causally. For window length $L$, define $L_{i,t}=\min(L,t+1)$ and the window ending at timestep $t$ as
\begin{equation}
    \mathcal{X}_{i,t}
    = [x_{i,t-L_{i,t}+1},\ldots,x_{i,t}].
    \label{eq:app_causal_window}
\end{equation}
Let $p_{i,\tau}^{(t)}$ be the classifier probability associated with position $\tau$ when evaluating $\mathcal{X}_{i,t}$. The causal window-mean score is
\begin{equation}
    q_i(t)
    = \frac{1}{L_{i,t}}
      \sum_{\tau=t-L_{i,t}+1}^{t}p_{i,\tau}^{(t)}.
    \label{eq:app_causal_score}
\end{equation}
When $L_{i,t}=L$, Eq.~\eqref{eq:app_causal_score} reduces to the window-mean aggregation in Eq.~\eqref{eq:episode_scores}. Evaluating every $t=0,\ldots,T_i$ produces the score curve $q_i(0:T_i)$. Sparse windows sampled for training or threshold-free episode scoring are not used to estimate alarm timing, because they do not represent the complete causal evaluation sequence. We enumerate the dense causal evaluations by $w=1,\ldots,W_i$, where $t_{i,w}$ is the endpoint of window $w$, and write $q_{\mathrm{mean},i}^{(w)}=q_i(t_{i,w})$. \figlink{fig:vla_fd_classifier} summarizes this aggregation stage.

\paragraph{Calibration masking and normalized time.}
The calibration-status feature $c_{i,t}$ identifies timesteps used for the episode-level GUARD calibration in Eq.~\eqref{eq:calibration_constant}. We suppress alarms while $c_{i,t}=1$ and define the first post-calibration timestep as
\begin{equation}
    t_i^{0}=1+\max\{t:c_{i,t}=1\}.
    \label{eq:app_post_calibration_start}
\end{equation}
The full rollout is retained for timing, but conformal calibration and alarm triggering use only $t\geq t_i^0$. Because rollout lengths differ, each post-calibration curve is linearly resampled onto a common grid $u_j=j/(J-1)$, $j=0,\ldots,J-1$, using the normalized time map
\begin{equation}
    u_i(t)=\frac{t-t_i^0}{T_i-t_i^0}.
    \label{eq:app_normalized_time}
\end{equation}
We denote the resulting resampled score function by $\widetilde q_i(u_j)$.

\paragraph{Functional conformal upper band.}
Let $\mathcal{D}_{\mathrm{cp}}^0$ contain $n$ successful seen-task calibration rollouts. At each normalized time $u_j$, we estimate the successful-rollout mean and scale,
\begin{equation}
    \mu(u_j)=\frac{1}{n}\sum_{i\in\mathcal{D}_{\mathrm{cp}}^0}
    \widetilde q_i(u_j),\qquad
    \sigma(u_j)=\sqrt{\frac{1}{n-1}\sum_{i\in\mathcal{D}_{\mathrm{cp}}^0}
    \bigl(\widetilde q_i(u_j)-\mu(u_j)\bigr)^2}.
    \label{eq:app_functional_moments}
\end{equation}
Each successful calibration rollout receives the one-sided functional nonconformity score
\begin{equation}
    A_i=\max_j
    \frac{\widetilde q_i(u_j)-\mu(u_j)}{\sigma(u_j)+\varepsilon}.
    \label{eq:app_functional_nonconformity}
\end{equation}
Let $A_{(1)}\leq\cdots\leq A_{(n)}$ be the ordered scores and $k_\alpha=\min\{n,\lceil(n+1)(1-\alpha)\rceil\}$. The conformal multiplier and upper band are
\begin{equation}
    \lambda_\alpha=A_{(k_\alpha)},\qquad
    B_\alpha(u_j)=\mu(u_j)+\lambda_\alpha\bigl(\sigma(u_j)+\varepsilon\bigr).
    \label{eq:app_functional_band}
\end{equation}
Unlike a single scalar threshold, $B_\alpha$ adapts to systematic changes in successful-rollout scores at early, middle, and late normalized times. The corresponding threshold for window $w$ of rollout $i$ is $\tau_{\alpha,i}(w)=B_\alpha(u_i(t_{i,w}))$.

\paragraph{Alarm rule and metrics.}
For a new rollout, the first alarm window is
\begin{equation}
    w_i^\star
    =\min\bigl\{w:t_{i,w}\geq t_i^0,\ 
    q_{\mathrm{mean},i}^{(w)}>\tau_{\alpha,i}(w)\bigr\},
    \label{eq:app_alarm_rule}
\end{equation}
with no alarm if the set is empty. For each $\alpha$, we report the true-positive rate (TPR), true-negative rate (TNR), and balanced accuracy $\mathrm{BalAcc}=(\mathrm{TPR}+\mathrm{TNR})/2$. Detection time is evaluated on failed rollouts as
\begin{equation}
    d_i=
    \begin{cases}
        w_i^\star/W_i, & \text{if an alarm occurs},\\
        1, & \text{otherwise},
    \end{cases}
    \label{eq:app_detection_time}
\end{equation}
so missed failures contribute an end-of-rollout detection rather than being omitted. Because $W_i$ counts dense causal evaluations over the complete rollout, including the masked calibration phase, $d_i$ preserves the full-rollout timing scale.

The level $\alpha$ controls the sensitivity--specificity trade-off illustrated in \figlink{fig:conformal_alpha_sensitivity}: smaller values produce a higher, more conservative band, whereas larger values generally produce more and earlier alarms. Under exchangeability between future successful rollouts and the successful conformal-calibration rollouts, the construction targets a successful-rollout crossing probability of at most $\alpha$ up to finite-sample and grid effects~\cite{vovk2005algorithmic,angelopoulos2021gentle}. For held-out tasks, where exchangeability with seen-task calibration trajectories may not hold, we interpret false-alarm control empirically through TNR and balanced accuracy.

\section{Additional Experiments}
\label{app:additional_experiments}

\subsection{Computational Overhead and Deployability}
\sectarget{app:computational_overhead}
\label{app:computational_overhead}

\tablink{tab:computational_overhead} decomposes GUARD's runtime cost by processing stage. Stage 1 performs the policy's normal action generation and the saliency backward pass used to identify influential cache entries. Stage 2 edits the selected KV-cache positions and evaluates the original and counterfactual conditionings as a batched single-step probe from the same noisy action state. The saliency backward pass accounts for most of the additional computation, while cache editing and probing contribute a smaller fraction.

For Pi0, GUARD increases the per-step cost from $4.379$ to $4.985$ TFLOPs, corresponding to a $1.14\times$ overhead. For SmolVLA, the cost increases from $0.640$ to $0.811$ TFLOPs, or $1.27\times$. Stage 3 is not shown because the lightweight temporal classifier operates on short windows of seven-dimensional diagnostics and consumes negligible FLOPs relative to the VLA forward, backward, and denoising operations. Thus, GUARD adds one lightweight saliency backward pass and one batched counterfactual probe without requiring multiple complete action-generation rollouts. These measurements quantify arithmetic cost; realized latency will additionally depend on hardware and implementation efficiency.

\begin{center}
\begin{minipage}{\textwidth}
\centering
\tabtarget{tab:computational_overhead}
\captionof{table}{Computational cost of GUARD per action-generation step. Stage 1 includes standard action generation and the saliency backward pass, while Stage 2 includes KV-cache editing and the counterfactual probe. The overhead ratio compares total action generation with and without GUARD.}
\label{tab:computational_overhead}
\begingroup
\scriptsize
\setlength{\tabcolsep}{3pt}
\renewcommand{\arraystretch}{1.22}
\begin{tabular}{>{\centering\arraybackslash}m{1.20cm}|*{5}{>{\centering\arraybackslash}m{2.00cm}|}>{\centering\arraybackslash}m{1.10cm}}
\toprule
\rowcolor{resultheadergray}
\textbf{Model}
& \multicolumn{2}{c|}{\textbf{Stage 1}}
& \multicolumn{1}{c|}{\textbf{Stage 2}}
& \multicolumn{3}{c}{\textbf{Total Cost}} \\
\rowcolor{resultheadergray}
& \shortstack{\textbf{Normal Action}\\\textbf{Generation}\\\textbf{(TFLOPs)}}
& \shortstack{\textbf{Saliency Backward}\\\textbf{Overhead}\\\textbf{(TFLOPs)}}
& \shortstack{\textbf{KV-Cache Edit}\\\textbf{+ Probe Overhead}\\\textbf{(TFLOPs)}}
& \shortstack{\textbf{Action Generation}\\\textbf{with GUARD}\\\textbf{(TFLOPs)}}
& \shortstack{\textbf{Action Generation}\\\textbf{without GUARD}\\\textbf{(TFLOPs)}}
& \shortstack{\textbf{Overhead}\\\textbf{($\times$)}} \\
\midrule
Pi0
& 4.379
& 0.451
& 0.155
& 4.985
& 4.379
& 1.14 \\
SmolVLA
& 0.640
& 0.125
& 0.046
& 0.811
& 0.640
& 1.27 \\
\bottomrule
\end{tabular}
\endgroup
\end{minipage}
\end{center}

\subsection{Leave-One-Feature-Out Analysis}
\sectarget{app:leave_one_feature_out}
\label{app:leave_one_feature_out}

To assess the contribution of each diagnostic, we retrain the temporal classifier seven times, each time removing one component of $x_t=[S_t,\theta_t,b_t,E_t,B_t,c_t,G_t]$, and compare the resulting seen- and unseen-task ROC-AUC with the full model.

\begin{center}
\begin{minipage}{\textwidth}
\centering
\figtarget{fig:leave_one_feature_out}
\includegraphics[width=0.96\textwidth]{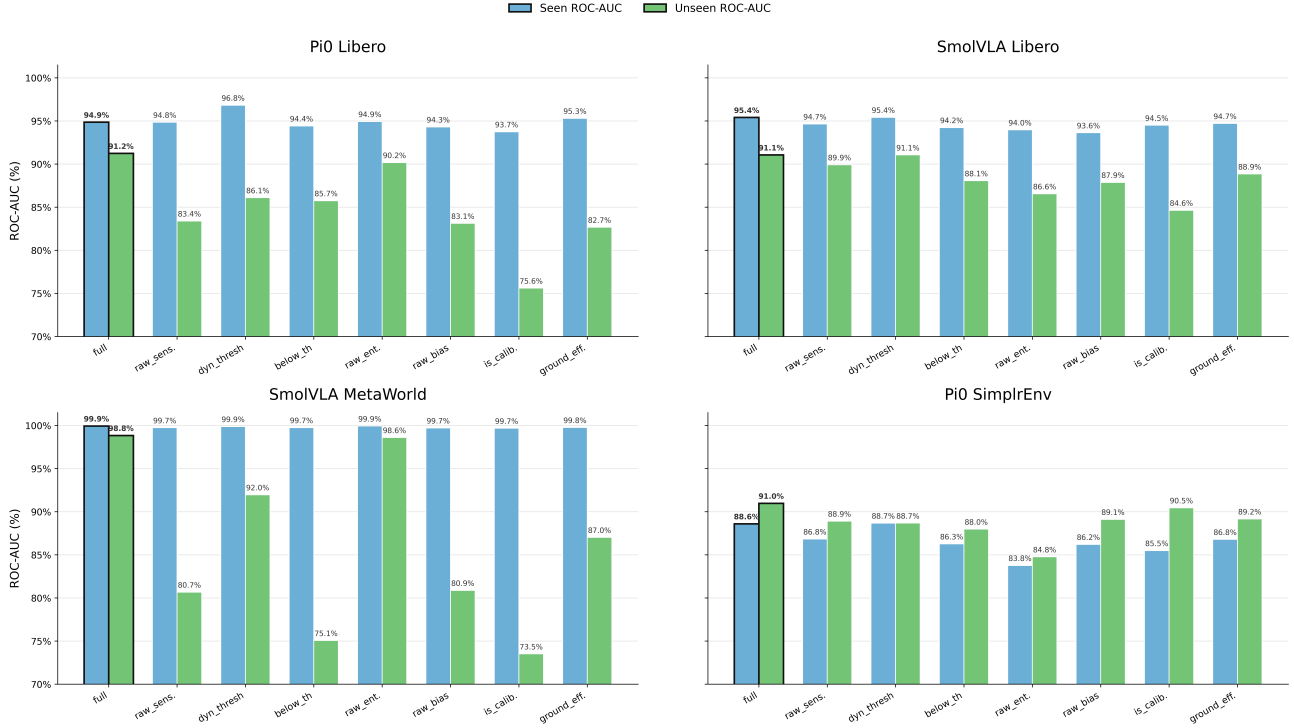}
\captionof{figure}{Leave-one-feature-out ROC-AUC analysis of the seven GUARD diagnostics. Each variant retrains the classifier after removing one feature; blue and green bars report seen- and unseen-task ROC-AUC, respectively. Larger reductions relative to the full model indicate a stronger contribution to failure classification and task-held-out generalization.}
\label{fig:leave_one_feature_out}
\end{minipage}
\end{center}

Seen-task ROC-AUC remains comparatively stable after most removals, whereas unseen-task performance is substantially more sensitive, indicating that the diagnostic diversity primarily supports transfer beyond the training tasks. The calibration-status feature $c_t$ is the most consistently important: removing it reduces unseen ROC-AUC from $91.2\%$ to $75.6\%$ on Pi0 LIBERO, from $91.1\%$ to $84.6\%$ on SmolVLA LIBERO, and from $98.8\%$ to $73.5\%$ on SmolVLA MetaWorld. MetaWorld additionally depends strongly on the below-threshold indicator $b_t$, raw sensitivity $S_t$, modality bias $B_t$, and grounding efficiency $G_t$, while Pi0 SimplerEnv shows its largest reduction when attention entropy $E_t$ is removed ($91.0\%$ to $84.8\%$). Although the dominant feature varies across policies and benchmarks, the full seven-feature representation achieves the best unseen-task ROC-AUC in every setting, showing that the diagnostics provide complementary grounding and calibration information.

\subsection{Choice of Saliency and Ablation Target}
\sectarget{app:saliency_target_ablation}
\label{app:saliency_target_ablation}

We compare the standard GUARD intervention on the final token-indexed KV cache with an alternative that operates on the input token embeddings. The two variants use the same rollout splits, top-$10\%$ modality-wise selection rule, diagnostic construction, temporal classifier, and evaluation protocol; they differ only in where saliency is measured and where the counterfactual intervention is applied. In the standard \emph{KV-cache} variant, the saliency backward pass differentiates the action-head objective with respect to the final visual- and language-token KV entries supplied directly as action-head conditioning, and the selected key and value entries are replaced by their modality-specific means. In the \emph{token-embedding} variant, gradients are propagated through both the action head and the VLM to the visual and language embeddings entering the VLM. The most salient input embeddings are then replaced by their modality-specific means, and the modified embeddings are propagated through the VLM to obtain the counterfactual conditioning used by the action head.

\begin{center}
\begin{minipage}{\textwidth}
\centering
\figtarget{fig:saliency_target_ablation}
\includegraphics[width=0.7\textwidth]{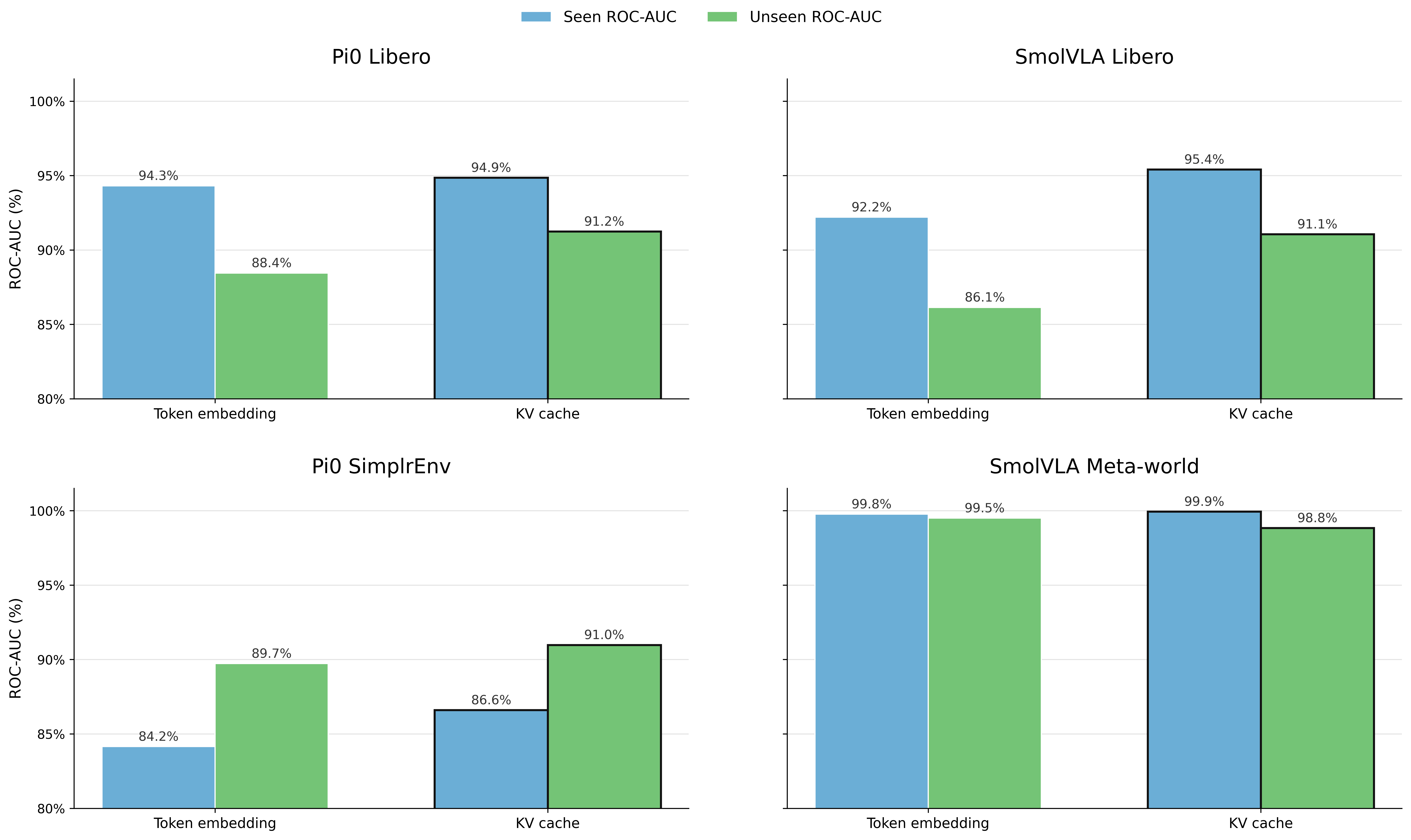}
\captionof{figure}{Saliency-target ablation comparing interventions on the input token embeddings and the final token-indexed KV cache. Blue and green bars report seen- and unseen-task ROC-AUC, respectively, across four manipulation policy--benchmark settings. All components other than the saliency and counterfactual-ablation target are held fixed.}
\label{fig:saliency_target_ablation}
\end{minipage}
\end{center}

\figlink{fig:saliency_target_ablation} shows that targeting the final KV cache improves seen-task ROC-AUC in all four settings and unseen-task ROC-AUC in three of four settings. The largest gain occurs on SmolVLA LIBERO, where the KV-cache target improves seen ROC-AUC from $92.2\%$ to $95.4\%$ and unseen ROC-AUC from $86.1\%$ to $91.1\%$. On Pi0 LIBERO, the unseen result increases from $88.4\%$ to $91.2\%$, while Pi0 SimplerEnv improves from $84.2\%$ to $86.6\%$ on seen tasks and from $89.7\%$ to $91.0\%$ on unseen tasks. SmolVLA MetaWorld is near saturation for both variants: the KV-cache intervention is slightly higher on seen tasks ($99.9\%$ versus $99.8\%$), whereas the token-embedding intervention is $0.7$ points higher on unseen tasks. Averaged across the four settings, targeting the KV cache improves seen and unseen ROC-AUC by approximately $1.6$ and $2.1$ percentage points, respectively.

This advantage is consistent with the location of the two representations in the action-generation computation. The final KV cache is the contextualized conditioning memory consumed directly by the action head, so its gradients more closely reflect the evidence governing the current denoising response, and ablating a selected cache entry produces a localized change at the action-head interface. Input token embeddings are computationally farther from the action output: their influence passes through multiple VLM layers, attention interactions, residual paths, normalization, and cross-modal mixing before reaching the action head. Saliency propagated to this earlier representation can therefore become more diffuse, while embedding-level ablations may be redistributed or partially compensated by the VLM. The results support the GUARD design choice of probing the final token-indexed KV cache as a more direct and discriminative representation of the multimodal evidence used for action generation.

\subsection{Sensitivity to the Conformal Miscoverage Level}
\sectarget{app:conformal_alpha_analysis}
\label{app:conformal_alpha_analysis}

\begin{center}
\begin{minipage}{\textwidth}
\centering
\figtarget{fig:conformal_alpha_sensitivity}
\includegraphics[width=0.96\textwidth]{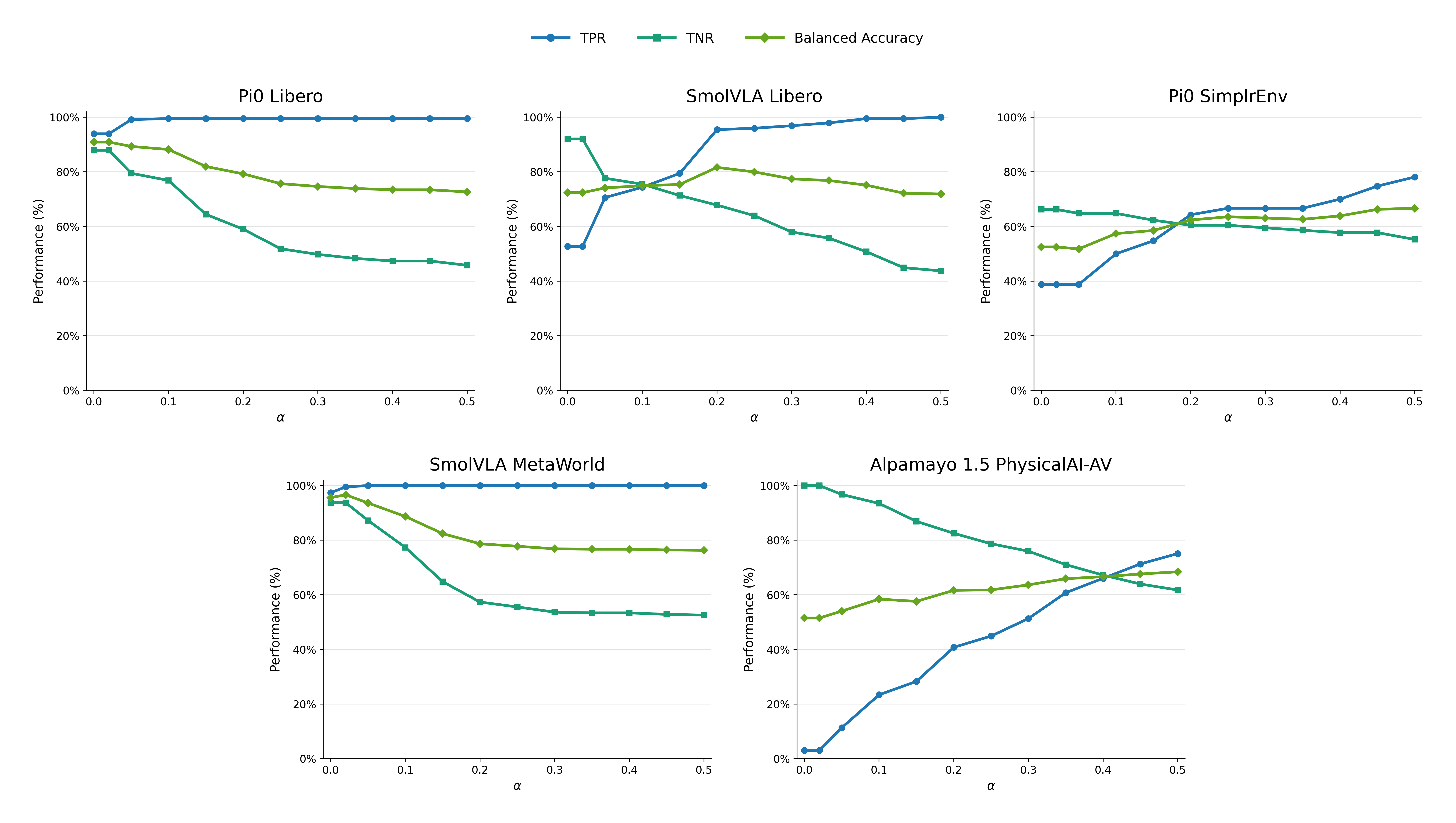}
\captionof{figure}{Effect of the functional conformal miscoverage level $\alpha$ on GUARD's unseen-task online failure detection across five policy--benchmark settings. Curves report the true-positive rate (TPR), true-negative rate (TNR), and balanced accuracy. Increasing $\alpha$ generally relaxes the alarm threshold, improving failure sensitivity while reducing specificity; balanced accuracy summarizes this trade-off.}
\label{fig:conformal_alpha_sensitivity}
\end{minipage}
\end{center}

\section{Implementation Details}
\sectarget{app:implementation_details}
\label{app:implementation_details}

\subsection{Evaluation Benchmarks}
\sectarget{app:evaluation_benchmarks}
\label{app:evaluation_benchmarks}

For each task-held-out split, complete tasks are reserved for unseen-task evaluation, while episodes from the remaining tasks are stratified by success/failure label into training and seen-task evaluation sets. All temporal samples from a rollout remain in the same partition, preventing leakage and measuring transfer to tasks without additional labeled rollouts.

\paragraph{Benchmark setup.}
\tablink{tab:benchmark_setup} summarizes the benchmark sources and the
episode-level split protocol used for each VLA--benchmark variant. For the
manipulation benchmarks, failure labels are derived from rollout success:
successful episodes are treated as negative examples for failure prediction,
whereas failed episodes are treated as positive examples. For Alpamayo, we use
the PhysicalAI-AV autonomous-driving dataset and define failures using the
minimum-ADE threshold of $3.0$. In all cases, splits are constructed at the
episode or clip level so that prefixes from the same rollout do not appear in
both training and evaluation subsets.

\paragraph{Aggregated split statistics.}
\tablink{tab:benchmark_split_statistics} reports the aggregate number of
episode or clip assignments across the three seen/unseen splits. The train
column corresponds to the training episodes from seen tasks, eval-seen
corresponds to held-out episodes from seen tasks, and eval-unseen corresponds
to all episodes from held-out tasks. Because each benchmark variant is evaluated
under three different seen/unseen splits, the total in this table counts split
assignments rather than unique original rollouts. Thus, a rollout can contribute
to different columns across different split folds if its task is seen in one
split and unseen in another.

\paragraph{LIBERO task statistics.}
\tablink{tab:libero_task_statistics} gives the per-task rollout outcomes for
the two LIBERO-10 variants. Each LIBERO-10 task contains 50 evaluated rollouts.
The table reports the number and percentage of successful rollouts for each
task before constructing the classifier splits. These values characterize the
base VLA rollout distribution from which the failure-classification dataset is
formed. The same LIBERO-10 task IDs and split protocol are used for the
Pi0 and SmolVLA variants, but their success rates differ because the
rollouts are generated by different policies.

\paragraph{SimplerEnv and Meta-World task statistics.}
\tablink{tab:simpler_metaworld_task_statistics} reports task-level success
statistics for the Pi0--SimplerEnv and SmolVLA--Meta-World variants. The
SimplerEnv benchmark uses four Google Robot tasks with 50 rollouts per task.
The Meta-World benchmark uses a 10-task subset of MT50, also with 50 rollouts
per task. These per-task success rates determine the class balance of the
failure-detection dataset and help identify tasks that provide more failure
examples for classifier training and evaluation.

\paragraph{Alpamayo / PhysicalAI-AV scenario statistics.}
\tablink{tab:alpamayo_task_statistics} reports the per-scenario clip counts
used for the Alpamayo failure-classification benchmark. Unlike the manipulation
benchmarks, these labels are not based on binary rollout success from a
simulator. Instead, each clip is labeled using a minimum-ADE threshold of
$3.0$: clips above the threshold are counted as failures, and clips below the
threshold are counted as successes. The local benchmark subset contains 584
unique clips across 12 scenario IDs, with 50 clips for most scenarios and fewer
clips for scenario IDs 6 and 11.

\begin{center}
\begin{minipage}{\textwidth}
\centering
\tabtarget{tab:benchmark_setup}
\captionof{table}{Benchmark setup for the five evaluated variants. Failure labels for manipulation benchmarks are derived from rollout success; Alpamayo failures are defined by a minimum-ADE threshold of $3.0$.}
\label{tab:benchmark_setup}
\begingroup
\small
\setlength{\tabcolsep}{4pt}
\renewcommand{\arraystretch}{1.15}
\begin{tabularx}{\textwidth}{
@{}>{\raggedright\arraybackslash}p{3.0cm}
   >{\raggedright\arraybackslash}p{3.1cm}
   >{\centering\arraybackslash}p{1.4cm}
   >{\centering\arraybackslash}p{1.8cm}
   Y@{}}
\toprule
Variant & Benchmark / dataset & Tasks & Rollouts & Split protocol \\
\midrule
Pi0--LIBERO & LIBERO-10 & 10 & 50 per task & 3 unseen tasks per split; remaining 7 tasks split 60/40 into train/eval-seen \\
SmolVLA--LIBERO & LIBERO-10 & 10 & 50 per task & Same LIBERO split protocol \\
Pi0--SimplerEnv & SimplerEnv Google Robot & 4 & 50 per task & 1 unseen task per split; remaining 3 tasks split 60/40 into train/eval-seen \\
SmolVLA--Meta-World & Meta-World MT10 subset & 10 & 50 per task & 3 unseen tasks per split; remaining 7 tasks split 60/40 into train/eval-seen \\
Alpamayo & PhysicalAI-AV & 12 & 584 total clips & 3 unseen scenario IDs per split; seen scenarios split 60/40 into train/eval-seen \\
\bottomrule
\end{tabularx}
\endgroup
\end{minipage}
\end{center}

\begin{center}
\begin{minipage}{\textwidth}
\centering
\tabtarget{tab:benchmark_split_statistics}
\captionof{table}{Aggregated benchmark split statistics across the three seen/unseen splits. Counts are episode or clip assignments across splits.}
\label{tab:benchmark_split_statistics}
\begingroup
\small
\setlength{\tabcolsep}{4pt}
\renewcommand{\arraystretch}{1.12}
\begin{tabularx}{\textwidth}{
@{}>{\raggedright\arraybackslash}p{3.0cm}
   >{\centering\arraybackslash}p{1.1cm}
   >{\centering\arraybackslash}p{1.4cm}
   >{\centering\arraybackslash}p{1.4cm}
   >{\centering\arraybackslash}p{1.4cm}
   >{\centering\arraybackslash}p{1.4cm}
   >{\centering\arraybackslash}p{1.4cm}
   Y@{}}
\toprule
Variant & Tasks & train & eval-seen & eval-unseen & Total & Success rate & Unseen task IDs by split \\
\midrule
Pi0--LIBERO & 10 & 630 & 420 & 450 & 1500 & 46.2\% & $\{0,3,5\}$; $\{5,7,8\}$; $\{6,8,9\}$ \\
SmolVLA--LIBERO & 10 & 630 & 420 & 450 & 1500 & 45.8\% & $\{0,3,5\}$; $\{5,7,8\}$; $\{6,8,9\}$ \\
Pi0--SimplerEnv & 4 & 270 & 180 & 150 & 600 & 64.0\% & $\{0\}$; $\{3\}$; $\{2\}$ \\
SmolVLA--Meta-World & 10 & 630 & 420 & 450 & 1500 & 69.8\% & $\{4,18,31\}$; $\{31,40,48\}$; $\{38,48,49\}$ \\
Alpamayo / PhysicalAI-AV & 12 & 783 & 521 & 448 & 1752 & 40.9\% & $\{8,9,11\}$; $\{1,8,11\}$; $\{2,7,8\}$ \\
\bottomrule
\end{tabularx}
\endgroup
\end{minipage}
\end{center}

\begin{center}
\begin{minipage}{\textwidth}
\centering
\tabtarget{tab:libero_task_statistics}
\captionof{table}{Per-task LIBERO-10 rollout success statistics for the two LIBERO variants.}
\label{tab:libero_task_statistics}
\begingroup
\scriptsize
\setlength{\tabcolsep}{3pt}
\renewcommand{\arraystretch}{1.12}
\begin{tabularx}{\textwidth}{
@{}>{\centering\arraybackslash}p{0.7cm}
   Y
   >{\centering\arraybackslash}p{1.4cm}
   >{\centering\arraybackslash}p{1.6cm}@{}}
\toprule
ID & LIBERO-10 instruction & Pi0 success & SmolVLA success \\
\midrule
0 & put the white mug on the left plate and put the yellow and white mug on the right plate & 16/50 (32\%) & 14/50 (28\%) \\
1 & put the white mug on the plate and put the chocolate pudding to the right of the plate & 26/50 (52\%) & 31/50 (62\%) \\
2 & put the yellow and white mug in the microwave and close it & 35/50 (70\%) & 31/50 (62\%) \\
3 & turn on the stove and put the moka pot on it & 38/50 (76\%) & 40/50 (80\%) \\
4 & put both the alphabet soup and the cream cheese box in the basket & 10/50 (20\%) & 4/50 (8\%) \\
5 & put both the alphabet soup and the tomato sauce in the basket & 30/50 (60\%) & 32/50 (64\%) \\
6 & put both moka pots on the stove & 17/50 (34\%) & 24/50 (48\%) \\
7 & put both the cream cheese box and the butter in the basket & 18/50 (36\%) & 16/50 (32\%) \\
8 & put the black bowl in the bottom drawer of the cabinet and close it & 14/50 (28\%) & 17/50 (34\%) \\
9 & pick up the book and place it in the back compartment of the caddy & 27/50 (54\%) & 20/50 (40\%) \\
\bottomrule
\end{tabularx}
\endgroup
\end{minipage}
\end{center}

\begin{center}
\begin{minipage}{\textwidth}
\centering
\tabtarget{tab:simpler_metaworld_task_statistics}
\captionof{table}{Per-task rollout success statistics for Pi0--SimplerEnv and SmolVLA--Meta-World.}
\label{tab:simpler_metaworld_task_statistics}
\begingroup
\small
\setlength{\tabcolsep}{4pt}
\renewcommand{\arraystretch}{1.12}
\begin{tabularx}{\textwidth}{
@{}>{\raggedright\arraybackslash}p{3.0cm}
   >{\centering\arraybackslash}p{0.8cm}
   Y
   >{\centering\arraybackslash}p{1.5cm}@{}}
\toprule
Variant & ID & Task & Success \\
\midrule
Pi0--SimplerEnv & 0 & google\_robot\_move\_near\_v0 & 36/50 (72\%) \\
Pi0--SimplerEnv & 1 & google\_robot\_open\_drawer & 29/50 (58\%) \\
Pi0--SimplerEnv & 2 & google\_robot\_close\_drawer & 40/50 (80\%) \\
Pi0--SimplerEnv & 3 & google\_robot\_place\_apple\_in\_closed\_top\_drawer & 23/50 (46\%) \\
\midrule
SmolVLA--Meta-World & 40 & reach-v3 & 10/50 (20\%) \\
SmolVLA--Meta-World & 38 & push-v3 & 13/50 (26\%) \\
SmolVLA--Meta-World & 31 & pick-place-v3 & 26/50 (52\%) \\
SmolVLA--Meta-World & 15 & door-open-v3 & 47/50 (94\%) \\
SmolVLA--Meta-World & 18 & drawer-open-v3 & 47/50 (94\%) \\
SmolVLA--Meta-World & 17 & drawer-close-v3 & 50/50 (100\%) \\
SmolVLA--Meta-World & 4 & button-press-topdown-v3 & 50/50 (100\%) \\
SmolVLA--Meta-World & 28 & peg-insert-side-v3 & 6/50 (12\%) \\
SmolVLA--Meta-World & 48 & window-open-v3 & 50/50 (100\%) \\
SmolVLA--Meta-World & 49 & window-close-v3 & 50/50 (100\%) \\
\bottomrule
\end{tabularx}
\endgroup
\end{minipage}
\end{center}

\begin{center}
\begin{minipage}{\textwidth}
\centering
\tabtarget{tab:alpamayo_task_statistics}
\captionof{table}{Per-scenario PhysicalAI-AV statistics used for Alpamayo. Failure labels are assigned using the min-ADE threshold of $3.0$.}
\label{tab:alpamayo_task_statistics}
\begingroup
\small
\setlength{\tabcolsep}{5pt}
\renewcommand{\arraystretch}{1.12}
\begin{tabular}{
@{}>{\centering\arraybackslash}p{1.6cm}
   >{\centering\arraybackslash}p{1.4cm}
   >{\centering\arraybackslash}p{1.5cm}
   >{\centering\arraybackslash}p{1.5cm}
   >{\centering\arraybackslash}p{2.0cm}@{}}
\toprule
Scenario ID & Clips & Success & Failure & Success rate \\
\midrule
0 & 50 & 19 & 31 & 38.0\% \\
1 & 50 & 16 & 34 & 32.0\% \\
2 & 50 & 22 & 28 & 44.0\% \\
3 & 50 & 20 & 30 & 40.0\% \\
4 & 50 & 15 & 35 & 30.0\% \\
5 & 50 & 30 & 20 & 60.0\% \\
6 & 35 & 15 & 20 & 42.9\% \\
7 & 50 & 22 & 28 & 44.0\% \\
8 & 50 & 17 & 33 & 34.0\% \\
9 & 50 & 16 & 34 & 32.0\% \\
10 & 50 & 19 & 31 & 38.0\% \\
11 & 49 & 28 & 21 & 57.1\% \\
\bottomrule
\end{tabular}
\endgroup
\end{minipage}
\end{center}

\subsection{Hyperparameter Search}
\label{app:hyperparameter_search}

We search over classifier architecture, optimization parameters, and temporal
windowing. The four manipulation benchmarks share
the model and training search space reported in
\tablink{tab:manipulation_model_search}, while their temporal window and
searches are benchmark-specific
(\tablink{tab:libero_search} and \tablink{tab:simpler_metaworld_search}).
A separate Transformer-specific search space is used for the autonomous-driving
experiments, as reported in \tablink{tab:alpamayo_search}.

We denote a temporal-window configuration by $(w,n)$, where $w$ is the number
of timesteps in each window and $n$ is the number of windows sampled from each
episode.

\begin{center}
\begin{minipage}{\textwidth}
\centering
\tabtarget{tab:manipulation_model_search}
\captionof{table}{Shared model and training hyperparameter search space for the
Pi0--LIBERO, SmolVLA--LIBERO, Pi0--SimplerEnv, and
SmolVLA--Meta-World experiments. A gradient-clipping value of $0$ denotes
that gradient clipping is disabled.}
\label{tab:manipulation_model_search}
\begingroup
\small
\setlength{\tabcolsep}{6pt}
\renewcommand{\arraystretch}{1.12}
\begin{tabularx}{\textwidth}{
    @{}>{\raggedright\arraybackslash}p{2.8cm}
       >{\raggedright\arraybackslash}p{3.7cm}
       Y@{}}
\toprule
Category & Hyperparameter & Values searched \\
\midrule
\multirow{3}{*}{Architecture}
    & Classifier model
    & \{GRU, LSTM, Transformer\} \\
    & Hidden dimension
    & $\{32,64,96,128\}$ \\
    & Number of layers
    & $\{1,2,3\}$ \\
\midrule
\multirow{5}{*}{Optimization}
    & Dropout
    & $\{0,0.05,0.10,0.20,0.30\}$ \\
    & Learning rate
    & $\{3{\times}10^{-4},5{\times}10^{-4},
       1{\times}10^{-3},2{\times}10^{-3},3{\times}10^{-3}\}$ \\
    & Weight decay
    & $\{0,10^{-5},10^{-4},10^{-3}\}$ \\
    & Batch size
    & $\{128,256,512\}$ \\
    & Gradient-clipping norm
    & $\{0,0.5,1,2\}$ \\
\bottomrule
\end{tabularx}
\endgroup
\end{minipage}
\end{center}

\tablink{tab:best_hyperparameters_all_variants} reports the best configuration selected for each VLA--benchmark variant after evaluating the model, optimization, and temporal-window choices defined in \tablink{tab:manipulation_model_search}, \tablink{tab:libero_search}, \tablink{tab:simpler_metaworld_search}, and \tablink{tab:alpamayo_search}. The selected classifier architecture and training parameters differ across variants, while the reported window configuration specifies the temporal context used to train and evaluate the final failure detector.

\begin{center}
\begin{minipage}{\textwidth}
\centering
\tabtarget{tab:libero_search}
\captionof{table}{Dataset-specific temporal-window search spaces for
the LIBERO experiments. Each pair denotes
$(\text{window size},\text{windows per episode})$.}
\label{tab:libero_search}
\begingroup
\small
\setlength{\tabcolsep}{5pt}
\renewcommand{\arraystretch}{1.18}
\begin{tabularx}{\textwidth}{
    @{}>{\raggedright\arraybackslash}p{4.0cm}Y@{}}
\toprule
VLA and benchmark & Window configurations \\
\midrule
Pi0--LIBERO
&
$\{(20,8)\}$
\\[2pt]

SmolVLA--LIBERO
&
$\{(20,8)\}$
\\
\bottomrule
\end{tabularx}
\endgroup
\end{minipage}
\end{center}

\begin{center}
\begin{minipage}{\textwidth}
\centering
\tabtarget{tab:simpler_metaworld_search}
\captionof{table}{Dataset-specific temporal-window search spaces for
the SimplerEnv and Meta-World experiments. Each pair denotes
$(\text{window size},\text{windows per episode})$.}
\label{tab:simpler_metaworld_search}
\begingroup
\small
\setlength{\tabcolsep}{5pt}
\renewcommand{\arraystretch}{1.18}
\begin{tabularx}{\textwidth}{
    @{}>{\raggedright\arraybackslash}p{4.0cm}Y@{}}
\toprule
VLA and benchmark & Window configurations \\
\midrule
Pi0--SimplerEnv
&
$\begin{aligned}
\{&(3,16),(5,12),(8,8),(10,8),(12,6),\\
  &(15,6),(18,5),(20,4),(20,8)\}
\end{aligned}$
\\[3pt]

SmolVLA--Meta-World
&
All pairs in
$\{3,4,5,6\}\times\{4,8,16,32,64\}$,
where the first and second sets respectively specify the window size and
number of windows per episode
\\
\bottomrule
\end{tabularx}
\endgroup
\end{minipage}
\end{center}

\begin{center}
\begin{minipage}{\textwidth}
\centering
\tabtarget{tab:alpamayo_search}
\captionof{table}{Hyperparameter search space for the Alpamayo autonomous-driving
experiments. The classifier backbone is fixed to a Transformer.}
\label{tab:alpamayo_search}
\begingroup
\small
\setlength{\tabcolsep}{6pt}
\renewcommand{\arraystretch}{1.12}
\begin{tabularx}{\textwidth}{
    @{}>{\raggedright\arraybackslash}p{2.8cm}
       >{\raggedright\arraybackslash}p{3.8cm}
       Y@{}}
\toprule
Category & Hyperparameter & Values searched \\
\midrule
\multirow{6}{*}{Architecture}
    & Classifier model
    & \{Transformer\} \\
    & Hidden dimension
    & $\{64,96,128\}$ \\
    & Classifier-head layers
    & $\{1\}$ \\
    & Attention heads
    & $\{4,8\}$ \\
    & Feed-forward dimension
    & $\{64,128,256\}$ \\
    & Transformer layers
    & $\{1,2,3\}$ \\
\midrule
\multirow{5}{*}{Optimization}
    & Dropout
    & $\{0.15,0.20,0.25,0.30\}$ \\
    & Learning rate
    & $\{3{\times}10^{-4},5{\times}10^{-4},
       7{\times}10^{-4},10^{-3}\}$ \\
    & Weight decay
    & $\{0,10^{-6},10^{-5},10^{-4}\}$ \\
    & Batch size
    & $\{128,256,512\}$ \\
    & Gradient-clipping norm
    & $\{0.5,1,2\}$ \\
\midrule
\multirow{3}{*}{Temporal windows}
    & Window size $w$
    & $\{3,5\}$ \\
    & Windows per episode $n$
    & $\{2,3,4,5\}$ \\
    & Constraint
    & $w\leq 5$ \\
\bottomrule
\end{tabularx}
\endgroup
\end{minipage}
\end{center}

\begin{center}
\begin{minipage}{\textwidth}
\centering
\tabtarget{tab:best_hyperparameters_all_variants}
\captionof{table}{Best hyperparameters selected by the search for each VLA--benchmark variant.}
\label{tab:best_hyperparameters_all_variants}
\begingroup
\small
\setlength{\tabcolsep}{4pt}
\renewcommand{\arraystretch}{1.14}
\begin{tabularx}{\textwidth}{
    @{}>{\raggedright\arraybackslash}p{3.4cm}
       >{\centering\arraybackslash}p{2.2cm}
       >{\centering\arraybackslash}p{2.2cm}
       >{\centering\arraybackslash}p{2.2cm}
       >{\centering\arraybackslash}p{2.4cm}
       Y@{}}
\toprule
Hyperparameter
& Pi0--LIBERO
& SmolVLA--LIBERO
& Pi0--SimplerEnv
& SmolVLA--Meta-World
& Alpamayo \\
\midrule
Classifier model
& LSTM
& GRU
& Transformer
& LSTM
& Transformer \\

Hidden dimension
& $64$
& $64$
& $128$
& $128$
& $64$ \\

Number of layers
& $1$
& $1$
& $2$
& $1$
& $1$ \\

Attention heads
& -- & -- & -- & -- & $4$ \\

Feed-forward dimension
& -- & -- & -- & -- & $128$ \\

Transformer layers
& -- & -- & -- & -- & $2$ \\

Dropout
& $0.20$
& $0.20$
& $0.10$
& $0.30$
& $0.20$ \\

Learning rate
& $1{\times}10^{-3}$
& $1{\times}10^{-3}$
& $2{\times}10^{-3}$
& $3{\times}10^{-3}$
& $5{\times}10^{-4}$ \\

Weight decay
& $10^{-5}$
& $10^{-4}$
& $10^{-4}$
& $\{0,10^{-5},10^{-4}\}$
& $0$ \\

Batch size
& $128$
& $256$
& $512$
& $128$
& $256$ \\

Gradient-clipping norm
& $1$
& $0$
& $2$
& $2$
& $2$ \\

Window configuration $(w,n)$
& $(20,8)$
& $(20,8)$
& $(5,12)$
& $(6,8)$
& $(5,3)$ \\
\bottomrule
\end{tabularx}
\endgroup
\end{minipage}
\end{center}

\tablink{tab:guard_methodology_hparams} separately summarizes the hyperparameters used to generate the GUARD diagnostic stream before temporal classification. These settings control the action horizon, modality-specific saliency fractions, probe-noise scale, entropy correction, safety margin, calibration interval, and autonomous-driving temporal sampling; they are distinct from the classifier and optimization choices reported in \tablink{tab:best_hyperparameters_all_variants}. Entries marked ``n/a'' are not applicable to the corresponding policy--benchmark setting.

\begin{center}
\begin{minipage}{\textwidth}
\centering
\tabtarget{tab:guard_methodology_hparams}
\captionof{table}{Hyperparameters related to GUARD diagnostic creation.}
\label{tab:guard_methodology_hparams}
\begingroup
\scriptsize
\setlength{\tabcolsep}{4pt}
\renewcommand{\arraystretch}{1.15}
\begin{tabular}{p{1.2cm} p{5.4cm} c c c c c}
\toprule
\textbf{Symbol} & \textbf{Methodology meaning} & \textbf{SmolVLA MT10} & \textbf{SmolVLA LIBERO} & \textbf{PI0 LIBERO} & \shortstack{\textbf{PI0}\\\textbf{SimplerEnv}} & \textbf{Alpamayo AV} \\
\midrule
$H$ & Action horizon / chunk length & 10 & 10 & 10 & 10 & n/a \\
$\rho_{\mathrm{vis}}$ & Visual saliency fraction & 0.1 & 0.1 & 0.1 & 0.1 & 0.1 \\
$\rho_{\mathrm{lang}}$ & Language saliency fraction & 0.1 & 0.1 & 0.1 & 0.1 & 0.1 \\
$\lambda_{\mathrm{p}}$ & Probe-noise mixing scale in Eq.~(8) & 0.05 & 0.05 & 0.05 & 0.05 & 0.05 \\
$\gamma$ & Entropy correction exponent in Eqs.~(13)--(14) & 2.0 & 2.0 & 2.0 & 2.0 & 2.0 \\
$\mu$ & Safety margin in adaptive threshold & 0.8 & 0.8 & 0.8 & 0.8 & 0.8 \\
$\mathcal{C}_{\mathrm{start}}$ & Calibration window start step & 25 & 25 & 25 & 4 & 0 \\
$\mathcal{C}_{\mathrm{end}}$ & Calibration window end step & 35 & 35 & 35 & 10 & 3 \\
Temporal sampling & PhysicalAI AV temporal offsets & n/a & n/a & n/a & n/a & $2.0 \times 10^6$, $1.0 \times 10^6$ $\mu$s \\
\bottomrule
\end{tabular}
\endgroup
\end{minipage}
\end{center}


\begin{center}
\begin{minipage}{\textwidth}
\centering
\tabtarget{tab:notation_summary}
\captionof{table}{Summary of the principal notation used throughout the paper and appendix.}
\label{tab:notation_summary}
\begingroup
\scriptsize
\setlength{\tabcolsep}{4pt}
\renewcommand{\arraystretch}{1.08}
\begin{tabularx}{\textwidth}{
    @{}>{\centering\arraybackslash}p{3.0cm}Y
       >{\centering\arraybackslash}p{3.0cm}Y@{}}
\toprule
Symbol & Name and meaning & Symbol & Name and meaning \\
\midrule
\multicolumn{4}{@{}l}{\textit{Policy, action, and KV-cache notation}} \\
$t$ & Control or action-generation timestep
& $o_t$ & Multimodal observation at timestep $t$ \\
$I_t,\ell_t,s_t$ & Camera input, language instruction, and optional robot state
& $\pi_\theta$ & Frozen pretrained VLA policy \\
$A_t$ & Predicted action chunk beginning at timestep $t$
& $H,d_a$ & Action horizon and action dimension \\
$Z_t$ & Final VLM KV cache conditioning the action head
& $z_i=(\mathbf{k}_i,\mathbf{v}_i)$ & Key--value cache entry at position $i$ \\
$N$ & Number of positions in the final KV cache
& $K$ & Number of diffusion or flow denoising steps \\
\addlinespace[2pt]
\multicolumn{4}{@{}l}{\textit{Saliency and counterfactual-probe notation}} \\
$\Phi(A_t)$ & Scalar action-level objective used for saliency
& $g_i$ & Gradient saliency of the KV-cache entry at position $i$ \\
$\mathcal{M}_{\mathrm{vis}},\mathcal{M}_{\mathrm{lang}}$ & Sets of visual and language KV-cache positions
& $\rho_{\mathrm{vis}},\rho_{\mathrm{lang}}$ & Modality-specific saliency fractions \\
$\mathcal{S}_{\mathrm{vis}},\mathcal{S}_{\mathrm{lang}},\mathcal{S}$ & Selected salient positions and their union
& $\bar z_m$ & Mean KV-cache representation for modality $m$ \\
$Z_t^{-\mathrm{vis}},Z_t^{-\mathrm{lang}},Z_t^{-\mathrm{both}}$ & Visual, language, and joint mean-ablated caches
& $\xi,\lambda_{\mathrm{p}}$ & Probe noise and probe-noise mixing scale \\
$\widetilde n_{K-1}$ & Shared noisy action input used by all probes
& $D_\theta$ & Frozen action-head denoising-response function \\
\shortstack{$r_t^{+},r_t^{-\mathrm{vis}},$\\$r_t^{-\mathrm{lang}},r_t^{-\mathrm{both}}$} & Original and modality-ablated denoising responses
& $\varepsilon$ & Positive numerical stabilizer \\
\addlinespace[2pt]
\multicolumn{4}{@{}l}{\textit{GUARD diagnostics and online calibration}} \\
$S_t$ & Counterfactual sensitivity to jointly ablated evidence
& $B_t$ & Visual-to-language modality-bias ratio \\
$\bar P_{q,i}$ & Head-averaged attention probability from query $q$ to cache position $i$
& $\mathcal{Q}$ & Set of action queries used for attention aggregation \\
$E_t$ & Mean cross-attention entropy
& $G_t$ & Grounding efficiency $S_t/(E_t+\varepsilon)$ \\
$\mathcal{C}$ & Initial episode-calibration timestep set
& $C$ & Episode-level entropy-adjusted sensitivity baseline \\
$\gamma$ & Entropy-correction exponent
& $\mu$ & Safety-margin multiplier for the adaptive threshold \\
$\theta_t$ & Adaptive sensitivity threshold
& $c_t$ & Indicator that episode-level calibration is active \\
$b_t$ & Low-sensitivity-event indicator
& $x_t$ & Seven-dimensional GUARD diagnostic vector \\
\addlinespace[2pt]
\multicolumn{4}{@{}l}{\textit{Temporal classification and functional conformal prediction}} \\
$L$ & Temporal-classifier window length
& $X_{s:s+L-1}$ & Fixed-length diagnostic window beginning at $s$ \\
$t_{\mathrm{LSE}}$ & First low-sensitivity-event timestep
& $y_t$ & Timestep-level classifier target \\
$i,T_i,y_i$ & Rollout index, final timestep, and trajectory-level label
& $\mathcal{X}_{i,t}$ & Causal diagnostic window ending at timestep $t$ \\
$p_{i,\tau}^{(t)}$ & Classifier probability at position $\tau$ within $\mathcal{X}_{i,t}$
& $q_i(t)$ & Causal window-mean failure score \\
$W_i,t_{i,w}$ & Number of dense windows and endpoint of window $w$
& $q_{\mathrm{mean},i}^{(w)}$ & Failure score assigned to dense window $w$ \\
$t_i^0$ & First post-calibration timestep
& $u_i(t),u_j$ & Rollout-normalized time and common grid point \\
$\widetilde q_i(u_j)$ & Resampled failure-score function
& $\mathcal{D}_{\mathrm{cp}}^0,n$ & Successful conformal-calibration set and its size \\
$\mu(u_j),\sigma(u_j)$ & Successful-rollout mean and scale functions
& $A_i$ & One-sided functional nonconformity score \\
$\alpha,k_\alpha$ & Miscoverage level and finite-sample quantile index
& $\lambda_\alpha$ & Conformal multiplier \\
$B_\alpha(u_j)$ & Functional conformal upper band
& $\tau_{\alpha,i}(w)$ & Time-varying alarm threshold for window $w$ \\
$w_i^\star$ & First window crossing the conformal band
& $d_i$ & Normalized failure-detection time \\
TPR, TNR, BalAcc & True-positive rate, true-negative rate, and balanced accuracy
& $m\in\{\mathrm{vis},\mathrm{lang}\}$ & Modality index \\
\bottomrule
\end{tabularx}
\endgroup
\end{minipage}
\end{center}

\end{document}